\documentclass[10pt,twocolumn,letterpaper]{article}

\usepackage[pagenumbers]{cvpr2025/cvpr} %

\usepackage{amsmath,amsfonts,bm}

\def\eqref#1{equation~\ref{#1}}

\def\1{\bm{1}}

\def\rt{{\textnormal{t}}}

\def\rx{{\textnormal{x}}}

\def\rvt{{\mathbf{t}}}

\def\rvx{{\mathbf{x}}}

\def\mI{{\bm{I}}}

\DeclareMathAlphabet{\mathsfit}{\encodingdefault}{\sfdefault}{m}{sl}
\SetMathAlphabet{\mathsfit}{bold}{\encodingdefault}{\sfdefault}{bx}{n}

\usepackage[utf8]{inputenc} %
\usepackage{amsmath}
\usepackage{url}            %
\usepackage{booktabs}       %
\usepackage{amsfonts}       %
\usepackage{nicefrac}       %
\usepackage{microtype}      %
\usepackage{xcolor}         %
\usepackage{ifthen}
\usepackage{xspace}  %
\usepackage{float}
\usepackage{graphicx}
\usepackage{gensymb}  %
\usepackage{subcaption}
\usepackage{multirow}
\usepackage{algorithm}
\usepackage{algpseudocode}
\usepackage{bm}  %
\usepackage{indentfirst}

\usepackage[accsupp]{axessibility}  %

\definecolor{cvprblue}{rgb}{0.21,0.49,0.74}
\usepackage[pagebackref,breaklinks,colorlinks,allcolors=cvprblue]{hyperref}

\def\paperID{28} %
\def\confName{CVPR}
\def\confYear{2025}

\title{Progressive Autoregressive Video Diffusion Models}

\author{
  Desai Xie\textsuperscript{† 1} \qquad Zhan Xu\textsuperscript{2} \qquad Yicong Hong\textsuperscript{2} \qquad Hao Tan\textsuperscript{2} \qquad Difan Liu\textsuperscript{2} \\
  Feng Liu\textsuperscript{2} \qquad Arie Kaufman\textsuperscript{1} \qquad Yang Zhou\textsuperscript{2}\\
  \\
  \textsuperscript{1}Stony Brook University \qquad \textsuperscript{2}Adobe Research
}

\newcommand\blfootnote[1]{%
  \begingroup
  \renewcommand\thefootnote{}\footnote{#1}%
  \addtocounter{footnote}{-1}%
  \endgroup
}

\begin{document}
\definecolor{TodoColor}{rgb}{0,0.5,0} 

\ifthenelse{\equal{1}{0}}  %
{
    \definecolor{TODOColor}{rgb}{0.976, 0.282, 0.235}
    \newcommand{\todos}[1]{{\color{TODOColor} \textbf{TODO}: #1}}
    
    \definecolor{DesaiColor}{rgb}{0.17, 0.8, 0.8}
    \newcommand{\desai}[1]{{\color{DesaiColor} \textbf{Desai}: #1}}
    
    \definecolor{HaoColor}{rgb}{0, 0.5, 0.88}
    \newcommand{\hao}[1]{{\color{HaoColor} \textbf{Hao}: #1}}
    
    \definecolor{YicongColor}{rgb}{0.976, 0.482, 0.725}
    \newcommand{\yicong}[1]{{\color{YicongColor} \textbf{Yicong}: #1}}
    
    \definecolor{DifanColor}{rgb}{0.486, 0.749, 0.482}
    \newcommand{\difan}[1]{{\color{DifanColor} \textbf{Difan}: #1}}
    
    \definecolor{ZhanColor}{rgb}{0.486, 0.0, 0.482}
    \newcommand{\zhan}[1]{{\color{ZhanColor} \textbf{Zhan}: #1}}
    
    \definecolor{FengColor}{rgb}{0.729, 0.482, 0.976}
    \newcommand{\feng}[1]{{\color{FengColor} \textbf{Feng}: #1}}
    
    \definecolor{YangColor}{rgb}{0.976, 0.725, 0.435}
    \newcommand{\yang}[1]{{\color{YangColor} \textbf{Yang}: #1}}
    
    \definecolor{ArieColor}{rgb}{0.4157, 0.3529, 0.8039}
    \newcommand{\arie}[1]{{\color{ArieColor} \textbf{Arie}: #1}}
}
{
    \newcommand{\todos}[1]{}
    \newcommand{\desai}[1]{}
    \newcommand{\hao}[1]{}
    \newcommand{\yicong}[1]{}
    \newcommand{\difan}[1]{}
    \newcommand{\zhan}[1]{}
    \newcommand{\feng}[1]{}
    \newcommand{\yang}[1]{}
    \newcommand{\arie}[1]{}
}

\newcommand{\ignore}[1]{}
\newcommand\numberthis{\addtocounter{equation}{1}\tag{\theequation}}

\def\opensora{\textit{O}\xspace}
\def\internalmodel{\textit{M}\xspace}
\def\suppl{Supplementary Material\xspace}
\def\appx{Appendix\xspace}

\maketitle

\begin{abstract}

Current frontier video diffusion models have demonstrated remarkable results at generating high-quality videos.
However, they can only generate short video clips, normally around 10 seconds or 240 frames, due to computation limitations during training.
Existing methods naively achieve autoregressive long video generation by directly placing the ending of the previous clip at the front of the attention window as conditioning, which leads to abrupt scene changes, unnatural motion, and error accumulation.
In this work, we introduce a more natural formulation of autoregressive long video generation by revisiting the noise level assumption in video diffusion models.
Our key idea is to \textbf{1.} assign the frames with per-frame, progressively increasing noise levels rather than a single noise level and \textbf{2.} denoise and shift the frames in small intervals rather than all at once.
This allows for smoother attention correspondence among frames with adjacent noise levels, larger overlaps between the attention windows, and better propagation of information from the earlier to the later frames.
\textbf{Video diffusion models} equipped with our \textbf{progressive} noise schedule can \textbf{autoregressively} generate long videos with much improved fidelity compared to the baselines and minimal quality degradation over time.
We present the first results on text-conditioned 60-second (1440 frames) long video generation at a quality close to frontier models.
Code and video results are available at \url{https://desaixie.github.io/pa-vdm/}.
\end{abstract}

\blfootnote{†This work is done while Desai is an intern at Adobe Research.}
\section{Introduction}

\begin{figure}[h]
    \centering
    \begin{minipage}{0.48\textwidth}
        \includegraphics[width=\linewidth]{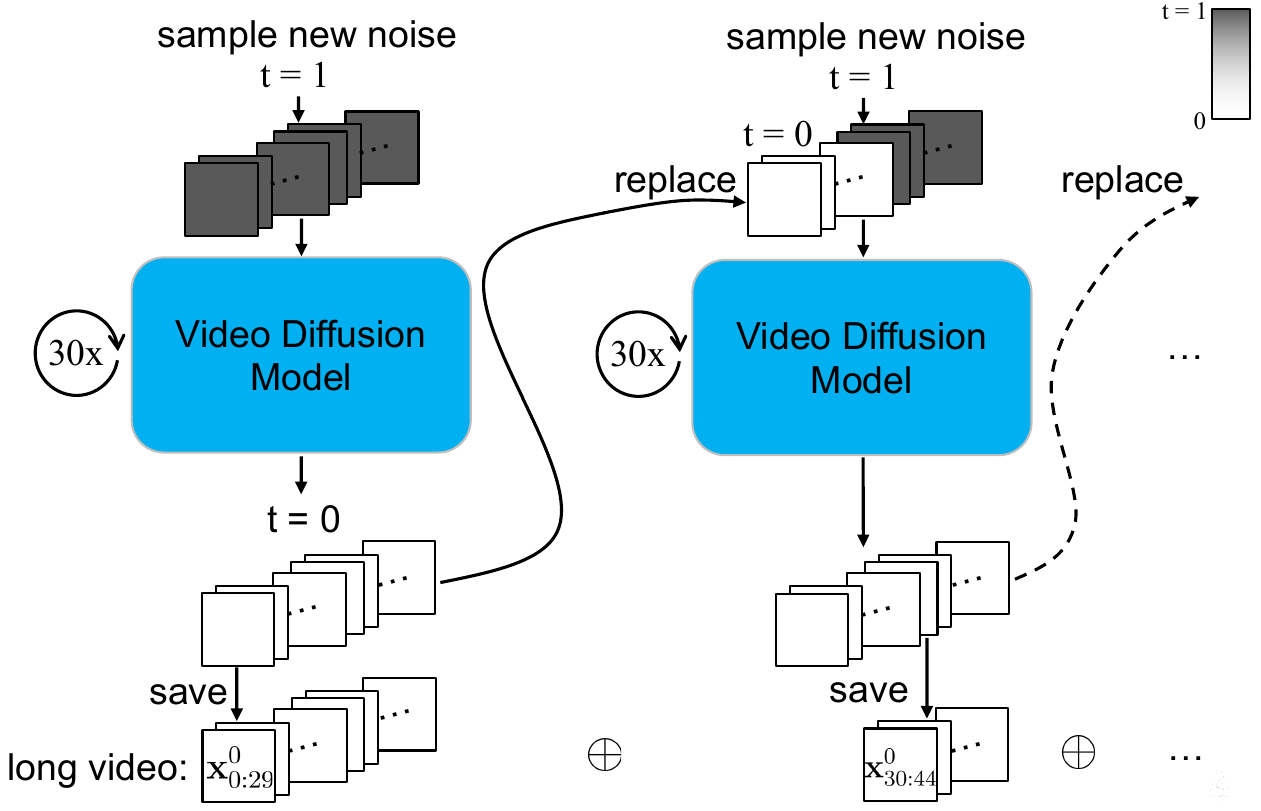}
        \subcaption{the replacement method}
    \end{minipage}\hfill
    \hspace{0.5em}\vline\hspace{0.5em}
    \begin{minipage}{0.48\textwidth}
        \includegraphics[width=\linewidth]{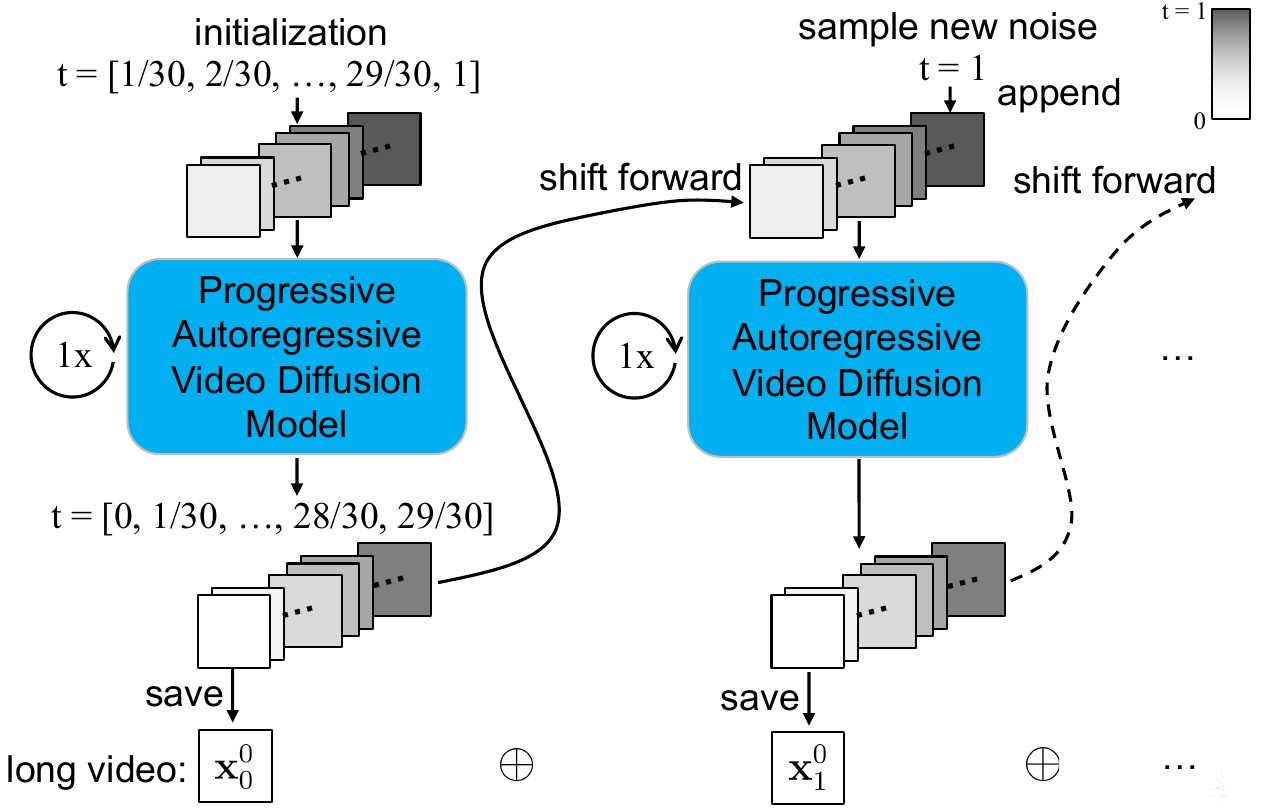}
        \subcaption{our PA-VDM}
    \end{minipage}\hfill
    
    \caption{
    Comparison of autoregressive long video generation methods.
    Top: the replacement method, which \textit{replaces} the front of noisy latent frames with the ending of previous clip as condition and denoising all the frames at once.
    Bottom: our PA-VDM, which applies \textit{progressive} noise levels and denoises and shifts the frames in small intervals.
    The final long video consists of autoregressively generated clean frames.
    $\oplus$ denotes concatenation.
    The noise level $t$ for each frame is illustrated by the solid color of the frame, where darker colors are closer to $1$ and lighter colors are closer to $0$.
    }
    \label{fig:main}
\end{figure}

Frontier video diffusion models~\cite{openai2024sora,meta2024moviegen,genmo_mochi1,yang2024cogvideox,runway_gen3,luma_dreammachine,kuaishou_kling,lin2024stivscalabletextimage,kong2025hunyuanvideosystematicframeworklarge,wanteam2025wanopenadvancedlargescale,seawead2025} have recently demonstrated remarkable success in generating high-quality video contents by scaling up transformer-based~\cite{peebles2023dit,NIPS2017_attention} architectures. 
However, they can only generate videos of relatively short duration, typically up to about 10 seconds or 240 frames, due to the demanding computation cost of long-sequence training. 
This temporal restriction leads to challenges for broader applications that require longer, more continuous video outputs.

Several approaches~\cite{ho2022videodiffusionmodels,henschel2024streamingt2vconsistentdynamicextendable,blattmann2023stable,opensora,gao2024vidgptintroducinggptstyleautoregressive} have been proposed to autoregressively apply video diffusion models for long video generation;
they generate short video clips in a windowed fashion, where each subsequent clip conditions on the final frames of the previous one. 
One solution~\cite{opensora,gao2024vidgptintroducinggptstyleautoregressive} directly places the conditioning frames into the input frames, replacing the noisy frames.
Another solution~\cite{song2020score,ho2022videodiffusionmodels} additionally adds the same level of noise to the conditioning frames as the noisy frames. 
This naive way of conditioning suffers from various flaws, including temporal inconsistency, abrupt scene changes, unnatural motion dynamics, and accumulated errors that lead to divergence.

In this work, we propose \textit{Progressive Autoregressive Video Diffusion Models} (PA-VDM) for high-quality long video generation.
The core innovation of our method lies in the denoising process: instead of applying a single noise level across all frames used in traditional video diffusion models~\cite{ho2022videodiffusionmodels,blattmann2023align}, we apply progressively increasing noise levels across the frames;
correspondingly, we denoise and shift the frames in small intervals, instead of denoising and shift them all at once. 
We illustrate our method in~\cref{fig:main}.
Such progressive noise levels and autoregressive video denoising benefit from larger overlaps between subsequent attention windows, smoother attention correspondence among frames with adjacent noise levels, and better propagation of information from the earlier to the later frames.
When applying our \textit{variable length} progressive noise schedule, our models can start or end the autoregressive generation at arbitrary video lengths.
Our \textit{chunked frames} and \textit{overlapped conditioning} techniques prevent divergent results and chunk-to-chunk discontinuity.
Together, our method can autoregressively generate long videos while maintaining the initial quality over time.

PA-VDM provides a range of benefits for the video generation community.
It can be easily implemented by changing the noise scheduling and finetuning pre-trained video diffusion models without changing the original model architecture;
this allows our method to be easily reproduced and combined with orthogonal methods, such as external memory modules~\cite{henschel2024streamingt2vconsistentdynamicextendable} and multiple text prompts~\cite{NEURIPS2024_StoryDiffusion,guo2025longcontexttuningvideo}.
While we choose to demonstrate PA-VDM on Diffusion Transformer (DiT)-based~\cite{peebles2023dit,ma2024latte,openai2024sora} models, PA-VDM is model agnostic and can be extended to UNet-based~\cite{ronneberger2015u,ho2022videodiffusionmodels} models.
As shown in~\cref{sec:exp:long}, our method can work training-free, if the model has been trained on varied noise levels~\cite{opensora}.
Moreover, the additional inference computational cost of PA-VDM is minimal without sacrificing any generation quality, as opposed to previous works~\cite{wang2023genlvideomultitextlongvideo,qiu2024freenoisetuningfreelongervideo,henschel2024streamingt2vconsistentdynamicextendable} that need to trade off quality for efficiency, making this approach more efficient for practical use in long video generation.

We compare our method to the baselines on a text-conditioned 60-second (1440 frames) long video generation benchmark consisting of 40 real videos and their captions.
Our quantitative results demonstrate that our results have overall the best quality across various dimensions and are the best at maintaining these metrics over the entire 60-second duration.
Qualitatively, our method substantially outperforms the baselines in terms of temporal consistency, motion dynamics, and maintaining quality over time.
In human evaluation, our models are also favored over various baseline models.
Our ablation studies demonstrate the effectiveness of our \textit{chunked frames} and \textit{overlapped conditioning} techniques at preventing cumulative error and temporal jittering, respectively.
By applying our method to two base models and outperforming their respective baselines, we confirm its universal applicability to existing video diffusion models.
We encourage readers to check out our project webpage for video results qualitatively comparing ours and the baselines. 
To facilitate future research, we also release our code based on Open-Sora~\cite{opensora}.

\noindent We summarize our contribution as follows:
\begin{enumerate}
    \item We propose a progressive noise level schedule, an autoregressive video denoising algorithm, and the chunked frames and overlapped conditioning techniques. Together, these enable high-quality long video generation building upon pre-trained video diffusion models.
    \item We are the first to achieve 60-second long video generation with quality that are close to frontier models, when compared at the same resolution. On our 60-second long video generation benchmark, we achieve superior VBench and FVD scores, majority preference in human evaluations, and strong qualitative results. This marks a significant step forward in generating longer videos, a dimension that has not been explored by recent frontier video diffusion models~\cite{meta2024moviegen,runway_gen3,luma_dreammachine,genmo_mochi1,yang2024cogvideox,kuaishou_kling}.
    \item Our method benefits the video generation research community in many ways, including easy implementation and reproduction, training-free application, minimal additional inference cost, and universal applicability on video diffusion models.
\end{enumerate}

\section{Background}
\subsection{Video Diffusion Models}
Diffusion models~\cite{sohl2015diffusion,ho2020ddpm} are generative models that learn to generate samples from a data distribution 
$\displaystyle q(\rvx^0)$ 
through an iterative denoising process.
During training, data samples are first corrupted using the forward diffusion process $\displaystyle q(\rvx^t|\rvx^0)$
\begin{align}
    \label{eq:forward_diffusion}
    \displaystyle q\left(\rvx^t \middle| \rvx^0\right)&=\mathcal{N}(\rvx^t; \sqrt{\alpha^t}\rvx^0, (1-\alpha^t)\mI)\\ %
    \label{eq:add_noise}
    \displaystyle \rvx^t &= \sqrt{\alpha^t}\rvx^0 + \sqrt{1-\alpha^t}\epsilon %
\end{align}
where $\displaystyle t\in [0, T)$ is the noise level or diffusion timestep, $\displaystyle \bm{\epsilon} \sim \mathcal{N}(\mathbf{0}, \mI)$ is the noise, and $\bm{\alpha}^{1:T}$ is the variance schedule. 
With those noisy data samples $\displaystyle \rvx^t$, diffusion models are trained to fit to the data distribution $\displaystyle q(\rvx^0)$ by maximizing the variational lower bound~\cite{kingma2013vae} of the log likelihood of $\displaystyle \rvx^0$, which can be simplified into a mean squared error loss~\cite{ho2020ddpm}
\begin{align}
    \label{eq:ddpm_loss}
    \displaystyle \mathcal{L}(\theta) = \left\| \epsilon - \epsilon_\theta(\rvx^t, t) \right\|^2 %
\end{align}
where $t$ is uniform between $0$ and $T$, $\displaystyle \bm{\epsilon} \sim \mathcal{N}(\mathbf{0}, \mI)$ and $\bm{\epsilon}_\theta$ is the noise predicted by the model with parameters $\theta$.

At sampling time, we consider the sampling noise level schedule $\displaystyle \bm{\tau} = \{\tau_0, \tau_1, ..., \tau_{S}\}$
, which is a monotonically increasing subset of $t \in [0, T)$ of length $S+1$~\cite{song2021ddim}.
Starting from $\displaystyle \rvx^{\tau_S} \sim \mathcal{N}( \mathbf{0}, \mI ), \tau_S=T$, the reverse denoising process is iteratively applied as
\begin{align}
    \label{eq:ddim_reverse}
    \displaystyle p_\theta\left(\rvx^{\tau_{i-1}} \middle| \rvx^{\tau_i}\right) = q_\sigma\left( \rvx^{\tau_{i-1}} \middle| \rvx^\tau, f_\theta(\rvx^t, t)\right)
\end{align}
where $\displaystyle \hat{\rvx}^0=\displaystyle f_\theta(\rvx^t, t)$ is the $\displaystyle \rvx^0$ predicted by the model and $\displaystyle f_\theta(\rvx^t, t)$ is the DDIM~\cite{song2021ddim} reverse process equation, which we omit for simplicity.
This gives us a sequence of samples $\displaystyle \rvx^T, \rvx^{\tau_{S-1}}, \ldots, \rvx^{\tau_1}, \rvx^0$, and the last sample $\displaystyle \rvx^0$ is the clean output result.

Latent video diffusion models~\cite{ho2022videodiffusionmodels,blattmann2023align} are diffusion models that models latent representations of video data, consisting of $F$ latent frames $\displaystyle \rvx_{0:F-1}=\{\rx_0, \rx_1, \ldots, \rx_{F-1}\}$.
The video latent frames are usually spatially and temporally~\cite{yu2024languagemodelbeatsdiffusion} compressed through a VAE~\cite{kingma2013vae}.
For simplicity, we refer to latent video diffusion models as video diffusion models and latent frames as frames.
The same forward process, reverse process, and loss (\cref{eq:forward_diffusion,eq:add_noise,eq:ddpm_loss,eq:ddim_reverse}) can be applied to model these video data by treating all the frames as one entity, ignoring the correlation among the frames. 
Recent video diffusion models~\cite{meta2024moviegen,opensora} have employed various diffusion model variants~\cite{liu2022rectifiedflow,liu2023instaflow,lipman2023flowmatchinggenerativemodeling} to improve training and inference efficiency as well as output quality.
Nevertheless, our method is compatible with any diffusion model variant as long as the model corrupts the data $\displaystyle \rvx^t$ at the same noise levels $t$.

\subsection{Autoregressive Long Video Generation via Replacement}
\label{sec:bg:long_replacement}
Video diffusion models can only generate short video clips, because they are only trained on videos with a limited length $F$ due to GPU memory limit.
When adapted to generating $L > F$ latent frames at sampling time, their generation quality substantially degrades~\cite{qiu2024freenoisetuningfreelongervideo}.
The straightforward solution is to autoregressively apply video diffusion models, generating each video clip while conditioning on the previous clip.
In this paper, we refer to the $F$ frames that the video diffusion model processes as the \textit{attention window}.

Given $E<F$ clean frames $\displaystyle \rvx_{0:E}^0$ as condition, there are two methods for autoregressively applying video diffusion models.
\cite{opensora,gao2024vidgptintroducinggptstyleautoregressive,blattmann2023stable} place the clean condition frames $\displaystyle \Bar{\rvx}_{0:E-1}^{0}$ directly at the front of the attention window, directly replacing the sampled frames $\displaystyle \rvx_{0:E-1}^{\tau_i}$ at each denoising step
\begin{align}
    \displaystyle p_\theta\left(\Bar{\rvx}_{0:E-1}^0, \rvx_{E:F-1}^{\tau_{i-1}} | \Bar{\rvx}_{0:E-1}^0, \rvx_{E:F-1}^{\tau_i}\right)
\end{align}
We will refer to this method as the \textit{replacement-without-noise} method.

\cite{song2020score,ho2022videodiffusionmodels} additionally add noise to the condition frames
\begin{align}
    \displaystyle p_\theta\left(\Bar{\rvx}_{0:E-1}^{\tau_{i-1}}, \rvx_{E:F-1}^{\tau_{i-1}} \middle| \Bar{\rvx}_{0:E-1}^{\tau_i}, \rvx_{E:F-1}^{\tau_i}\right)
\end{align}
where $\displaystyle \Bar{\rvx}_{0:E-1}^{\tau_i}$ are the condition frames $\displaystyle \Bar{\rvx}_{0:E-1}^{0}$ noised via the forward process (\cref{eq:forward_diffusion,eq:add_noise}).
This maintains the same noise level distribution and training objective as regular video diffusion models.
We will refer to this method as the \textit{replacement-with-noise} method.
Note that~\cite{ho2022videodiffusionmodels} proposes reconstruction guidance for the \textit{replacement-with-noise} method but is not widely adopted.

Both the \textit{replacement-with-noise} method and the \textit{replacement-without-noise} method allow a video diffusion model to autoregressively generate video frames by conditioning on previous frames.
We consider them as baselines in our experiments in~\cref{sec:exp:long}.

See~\cref{sec:concurrent} for a detailed discussion of two parallel works~\cite{ruhe2024rolling,kim2024fifo} that share a high-level idea similar to our work.
Please refer to~\cref{sec:related} for related works.

\begin{figure}[tbp]
    \centering
    \begin{minipage}{0.24\textwidth}
        \includegraphics[width=\linewidth]{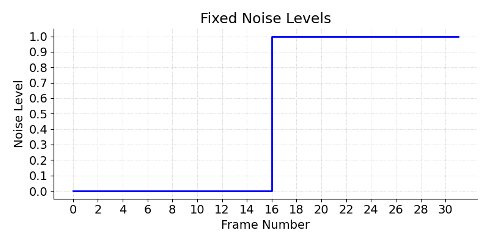}
    \end{minipage}\hfill
    \begin{minipage}{0.23\textwidth}
        \includegraphics[width=\linewidth]{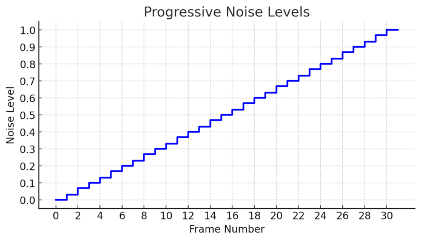}
    \end{minipage}\hfill
    
    \caption{
    Comparison of noise levels of a sequence of video frames when using the \textit{replacement without noise} method (left) and ours (right).
    }
    \label{fig:noise_level_comparison}
\end{figure}

\section{Progressive Autoregressive Video Diffusion Models}
\label{sec:method}

\begin{figure*}[htbp]
    \centering
    \begin{minipage}{.9\textwidth}

\begin{algorithm}[H]
\caption{Inference procedure of progressive autoregressive video diffusion models}\label{alg:main}
\begin{algorithmic}[1]

\Require Initial video latent frames $\displaystyle \rvx_{0:F-1}^0 = \{\rx_0^0, \rx_1^0, ..., \rx_{F-1}^0\}$, maximum noise level $T$, number of inference steps $S$, and attention window size $F=S$

\State $\displaystyle \bm{\tau}_{0:S} = \left\{\tau_0, \tau_1, \ldots, \tau_S \right\} = \left\{ 0, \frac{T}{S}, \ldots, T\right\}$ \Comment{\cref{eq:linear_schedule}, linear sampling noise level schedule}

\State $\displaystyle \bm{\epsilon} \sim \mathcal{N}( \mathbf{0} , \mI )$

\State $\displaystyle \rvx^{\bm{\tau}_{1:S}}_{0:F-1} = \sqrt{\alpha^{\bm{\tau}_{1:S}}}\rvx^0_{0:F-1} + \sqrt{1-\alpha^{\bm{\tau}_{1:S}}}\bm{\epsilon}$ \Comment{\cref{eq:add_noise}, add noise and set to progressive noise levels}

\For{each autoregressive generation step $\displaystyle i = 1, 2, \ldots, N$}
    \State $\displaystyle \rvx^{\bm{\tau}_{0:S-1}}_{0:F-1} = \left\{ \rx_0^0, \rx_1^{\tau_1}, \ldots, \rx_{F-1}^{\tau_{S-1}} \right\} 
    \sim p_\theta\left(\rvx^{\bm{\tau}_{0:S-1}}_{0:F-1} \middle| \rvx^{\bm{\tau}_{1:S}}_{0:F-1} \right)$ \Comment{\cref{eq:progressive}, one sampling step}
    
    \State $\displaystyle \rx_{F-1}^T \sim \mathcal{N}( \mathbf{0} , \mI )$ \Comment{Sample a new noisy frame}
    
    \State Append $\rx_0^0$ to the list of clean frames
    
    \State $\displaystyle \rvx^{\bm{\tau}_{0:S}}_{0:F-1} = \left\{ \rx_1^{\tau_1}, \ldots, \rx_{F-2}^{\tau_{S-1}}, \rx_{F-1}^{T} \right\}$ \Comment{Remove $\rx_0^0$, shift frames forward, and append $\rx_{F-1}^T$}

\EndFor

\State \Return List of clean frames

\end{algorithmic}
\end{algorithm}

    \end{minipage}
\end{figure*}

We consider long video generation with video diffusion models.
As discussed in~\cref{sec:bg:long_replacement}, existing video diffusion models can only generate short video clips up to a limited length $F$, and the replacement methods~\cite{ho2022videodiffusionmodels,opensora,gao2024vidgptintroducinggptstyleautoregressive} suffer from various flaws.
We describe a more natural formulation of autoregressive long video generation, which we call \textit{Progressive Autoregressive Video Diffusion Models} (PA-VDM).
We propose a per-frame progressively increasing noise schedule, which is inspired by~\cite{chen2024diffusionforcingnexttokenprediction}.
During training, we finetune pre-trained video diffusion models to adapt to our noise schedule;
during sampling, our models adopt such noise schedule and autoregressively generate video frames.

\subsection{Progressive Noise Levels and Autoregressive Generation}
\label{sec:method:progressive}
Conventional video diffusion methods assign a single noise level $t$ to all the latent frames.
Inspired by~\cite{chen2024diffusionforcingnexttokenprediction}, we adopt per-frame noise levels $\displaystyle \rvt_{0:F-1} = \{t_0, t_1, ..., t_{F-1}\}$ to the $F$ latent frames in the attention window.
In particular, we consider monotonically increasing noise levels for each frame,
where earlier frames are less noisy and later frames are more noisy.
In this work, we consider the linear sampling noise schedule with $S$ sampling steps
\begin{align}
    \label{eq:linear_schedule}
    \displaystyle \bm{\tau}_{0:S} = \left\{ 0, \frac{T}{S}, \frac{2T}{S}, \ldots, \frac{(S-1)T}{S}, T\right\}
\end{align}
which is monotonically increasing.
Given a sampling noise schedule,
instead of all the frames sharing a noise level and jointly going through the schedule as in conventional video diffusion models, each frame now goes through the schedule independently; at each step, the per-frame noise levels $\displaystyle \bm{\tau}$ still maintain the progressively increasing pattern.

Since both the sampling noise schedule and our target per-frame noise levels are monotonically increasing, we can now set per-frame noise levels $\displaystyle \rvt_{0:F-1}$ to be an interpolation of the sampling noise schedule $\displaystyle \bm{\tau}$.
Let us first consider the simple case of $F=S$, when our per-frame progressive noise levels can equal to either $\displaystyle \rvt = \bm{\tau}_{0:S-1}$ or $\displaystyle \rvt = \bm{\tau}_{1:S}$.
At each sampling step, the video diffusion model takes $\displaystyle \bm{\tau}_{0:S-1}$ as input and predicts $\displaystyle \bm{\tau}_{1:S}$
\begin{align}
    \label{eq:progressive}
    \displaystyle p_\theta\left(\rx_{0}^{\tau_0}, \rx_{1}^{\tau_1}, ..., \rx_{F-2}^{\tau_{S-2}}, \rx_{F-1}^{\tau_{S-1}} \middle| \rx_{0}^{\tau_1}, \rx_{1}^{\tau_2}, ..., \rx_{F-2}^{\tau_{S-1}}, \rx_{F-1}^{\tau_S}\right)
\end{align}
We illustrate progressive noise levels when $F=S$ in~\cref{fig:noise_level_comparison}.

Now we construct our autoregressive generation algorithm for video latent frames with progressive noise levels.
Notice that the input and output noise levels in~\cref{eq:progressive}, $\displaystyle \bm{\tau}_{0:S-1}$ and $\displaystyle \bm{\tau}_{1:S}$, only differ by $\displaystyle \tau_0=0$ and $\displaystyle \tau_S=T$.
We can simply transition the output frames back into the correct input noise levels by removing the clean frame $\displaystyle \rx_0^0$ at the front, shifting the frame sequence forward by one frame, and appending a new noisy frame $\displaystyle \rx_{F-1}^T \sim \mathcal{N}( \mathbf{0} , \mI )$ at back, as illustrated in~\cref{fig:main}.
We describe the autoregressive generation algorithm when $F=S$ in~\cref{alg:main}. 
The algorithm requires a clean short video $\displaystyle \bm{x}_{0:F-1}^0$ as initialization and extends from it.
We describe how to avoid this requirement in~\cref{sec:method:variable_length}.

More generally, when $F$ is a multiple of $S$, e.g. $F=90, S=30$, every set of $F/S=3$ frames would always share the same noise level during denoising and be removed from the attention window together as they reach $t=0$;
when $S$ is a multiple of $F$, e.g. $F=10, S=30$, the save, shift, and append operations for the sequence of frames (line 6, 7, 8 in~\cref{alg:main}) would happen once every $S/F=3$ steps.

Note that, regardless of what the noise level a frame initially has, it always goes through the same noise level schedule $\displaystyle \bm{\tau}_{S:0}$ as in conventional diffusion models.
Thus, for each individual frame, it is still modeled under the valid assumptions in diffusion model training~\cite{ho2020ddpm,lipman2023flowmatchinggenerativemodeling,liu2022rectifiedflow,liu2023instaflow} and sampling~\cite{song2021ddim}.
We only diverge from the noise level assumption in conventional video diffusion models~\cite{ho2022videodiffusionmodels}: 
now, each frame is modeled independently instead of jointly with the whole sequence of frames, and the progressive autoregressive video diffusion model attends to frames with different noise levels $\displaystyle \rvt_{0:F-1}$ instead of the same noise level $\displaystyle t$.
Thus, we can obtain our progressive autoregressive video diffusion models from pre-trained video diffusion models by adapting the model to the new noise level distribution through finetuning.
This saves us from the highly demanding computation cost of video diffusion model pre-training~\cite{meta2024moviegen,yang2024cogvideox}.

Intuitively, the benefit of our progressive video denoising process is that it gradually establishes correlation among consecutive latent frames.
Given some existing video frames as conditioning, it is challenging for video diffusion models to produce temporally consistent extension frames from newly sampled noisy frames~\cite{qiu2024freenoisetuningfreelongervideo}.
In contrast to the \textit{replacement-with-noise} method~\cite{blattmann2023stable,ho2022videodiffusionmodels} where the frames are denoised together at the same noise level, our progressive video denoising encourages the later frames with higher uncertainty to follow the patterns of the earlier and more certain frames, facilitating modeling a smoother temporal transition and better preserving motion velocity.
Compared to the \textit{replacement-without-noise} method where there is a large noise level gap between the clean condition frames $\displaystyle \Bar{\rvx}_{0:E-1}^{0}$ and the noisy frames $\displaystyle \rvx_{E:F-1}^{\tau_i}$, our method provides smoother attention correspondence, where the difference between neighboring noise levels is only $\frac{T}{S}$, as illustrated in~\cref{eq:linear_schedule,fig:noise_level_comparison}.

\subsection{Variable Length} 
\label{sec:method:variable_length}
The above design only allows for autoregressive video extension given an initial video of length $F$.
In addition, the noisy frames remaining in the attention window $\displaystyle \rvx_{0:F-1}^{\bm{\tau}_{1:S}}$  (line 8 of~\cref{alg:main}) are discarded after the end of the autoregressive inference, which can cause wasted computing resources and inaccurate handling of the ending of text prompt.
To enable text-to-long-video generation without any starting condition frames and properly ending the generation without wasting computation, we extend the base design in~\cref{eq:progressive,alg:main} to add an initialization stage and a termination stage, where the model operates on variable attention window lengths from $1$ to $F-1$.
During initialization, we simply disable the ``removing $\displaystyle \rx_0^0$'' operation in line 8 of~\cref{alg:main}:
starting from a noisy frame $\displaystyle \{\rx_0^T\}$, we denoise and append to obtain $\displaystyle \{ \rx_0^{\tau_{S-1}}, \rx_1^T \}$; we repeat this by $F-1$ times to obtain $\displaystyle \rvx^{\bm{\tau}_{1:S}}_{0:F-1} = \left\{ \rx_1^{\tau_1}, \ldots, \rx_{F-2}^{\tau_{S-1}}, \rx_{F-1}^{T} \right\}$, i.e. the input to line 5 of~\cref{alg:main}.
During termination, we disable the ``append $\displaystyle \rx_{F-1}^T$'' operation in line 6 and 7 of~\cref{alg:main}:
starting with $F$ frames $\displaystyle \rvx^{\bm{\tau}_{1:S}}_{0:F-1} = \left\{ \rx_0^{\tau_1}, \ldots, \rx_{F-2}^{\tau_{S-1}}, \rx_{F-1}^{T} \right\}$, we denoise, save and remove to obtain $\displaystyle \rvx^{\bm{\tau}_{1:S-1}}_{0:F-2} = \left\{ \rx_0^{\tau_1}, \ldots, \rx_{F-2}^{\tau_{S-1}} \right\}$; we repeat this by $F$ times to save and remove all the remaining frames in the attention window.
We train the model accordingly on video latent frames with variable lengths ranging from $1$ to $F-1$, following the noise levels described above. 

\subsection{Chunked Frames}
\label{sec:method:chunked_latents}
3D VAEs~\cite{kingma2013vae,yu2024languagemodelbeatsdiffusion,meta2024moviegen,opensora} usually encode and decode video latent frames chunk-by-chunk.
In our early experiments, we find that naively implementing our method on latent video diffusion models, i.e. when all latent frames are given different noise levels and the attention window is shifted by one frame at a time, leads to serious cumulative error and the videos diverge quickly after a few seconds, as shown in Ablation 2 in~\cref{fig:vs_ablation}.
We resolve the problem by \textit{treating a chunk of latent frames as a whole}: they are assigned with the same noise level, and are added and removed from the attention window together.
In other words, for a 3D VAE chunk size of $C$ latent frames, e.g. $C=5$ as mentioned in~\cref{sec:4.1}, we shift the attention window by $C$ frames every $C$ sampling steps.
Effectively, the $C$ frames that belong to the same chunk always have the same noise level $t$ and are added to or removed from the attention window together.
Our ablation experiments shows that, for models using a 3D VAE, treating a chunk of frames as a whole effectively prevents accumulated errors that would lead to divergence.

\subsection{Overlapped Conditioning}
\label{sec:method:overlapped_conditioning}
In our early experiments, naively implementing our method on video diffusion models results in temporal jittering.
We hypothesize that this is because the clean frames $\displaystyle \rvx_{0:C-1}^0$ are immediately removed from the attention window; 
as the later frames cannot attend to the previous clean frames, it is hard for the model to denoise the later frames to be perfectly temporally consistent with the previous clean frames.
In practice, we always keep a chunk of $C$ clean frames by prepending it to the attention window.
Our ablation study shows that overlapped conditioning helps resolving the frame-to-frame discontinuity issue.

\textit{Overlapped conditioning} requires an additional inference cost at $C/F$ ($5/50$ in our implementation) of the original cost.
When using the same number of conditioning frames $E$ and $F$, the replacement methods~\cite{ho2022videodiffusionmodels,opensora,gao2024vidgptintroducinggptstyleautoregressive} and ours have the same inference efficiency.
The key advantage of our method is that the large overlap of noisy frames enables the model to preserve the high-level information---such as motion---from prior frames.
Thus, we only need a single chunk of $C$ clean condition frames to propagate high-frequency details and prevent per-chunk temporal jittering. 
In contrast, the replacement methods need to balance the tradeoff between more overlap between video clips or better inference efficiency. 
In practice, their implementation~\cite{opensora} often use one chunk of frames as condition to save inference computation, but the limited overlap causes unnatural motion transition and abrupt scene changes across clips, as discussed in~\cref{sec:exp:long}.

\subsection{Training}
\label{sec:method:training}
As described in~\cref{sec:method:progressive}, PA-VDM requires change in the noise level distribution. We finetune pre-trained video diffusion model to adapt to our progressive noise level distribution.
Conventional diffusion model training~\cite{ho2020ddpm,liu2022rectifiedflow,liu2023instaflow} involves uniformly sampling a noise level $\rt\in [0, T)$, adding noise to the samples $\displaystyle \rvx^0_{0:F-1}$ via the forward diffusion process (\cref{eq:forward_diffusion,eq:add_noise}), and computing the loss (\cref{eq:ddpm_loss}).
During the finetuning process for PA-VDM, we simply continue with the conventional video diffusion model training but with our per-frame progressive training noise levels $\displaystyle \rvt_{0:F-1}$. 
In our experiment, we observed that, similar to the sampling noise levels $\displaystyle \bm{\tau}_{0:S}$ in~\cref{eq:linear_schedule}, training on a simple linear noise schedule yielded satisfactory results for all reported experiments. 
During training, the noise levels $\displaystyle \rvt$ is perturbated by a random shift $\displaystyle \delta$ to fully cover of the diffusion timestep range $[0, T)$~\cite{song2020denoising}. $\displaystyle \delta=0.4\epsilon(\rt_i-\rt_{i+1}), \bm{\epsilon} \sim \mathcal{N}( 0 , \mI )$ is randomly sampled for each training iteration and remains constant for all $\displaystyle \rvt_{0:F-1}$ within that iteration.

\section{Experiments}
\label{sec:4.1}

\begin{table*} [h]
\centering
\caption{
Quantitative comparison of our progressive autoregressive video generation (PA) and two baseline methods \textit{replacement-with-noise} (RW) and \textit{replacement-without-noise} (RN) on two base models (\internalmodel and \opensora), and other baselines StreamingT2V~\citep{henschel2024streamingt2vconsistentdynamicextendable}, Stable Video Diffusion (SVD)~\citep{blattmann2023stable}, and FIFO-Diffusion~\cite{kim2024fifo}.
}
\label{tab:main}
\setlength{\tabcolsep}{1pt}

\begin{tabular}{lcccccccc}
    \toprule
    & \multicolumn{1}{c}{Subject} & \multicolumn{1}{c}{Background} & \multicolumn{1}{c}{Motion} & \multicolumn{1}{c}{Dynamic} & \multicolumn{1}{c}{Aesthetic} & \multicolumn{1}{c}{Imaging} & \multicolumn{1}{c}{Num} & \multicolumn{1}{c}{}\\
    & \multicolumn{1}{c}{Consistency $\uparrow$} & \multicolumn{1}{c}{Consistency $\uparrow$} & \multicolumn{1}{c}{Smoothness $\uparrow$} & \multicolumn{1}{c}{Degree $\uparrow$} & \multicolumn{1}{c}{Quality $\uparrow$} & \multicolumn{1}{c}{Quality $\uparrow$} & \multicolumn{1}{c}{Scenes $\downarrow$} & \multicolumn{1}{c}{FVD $\downarrow$} \\
    \midrule
    \textbf{PA-\internalmodel} (ours) & 0.7923 & \textbf{0.8964} & 0.9896 & 0.8000 & \textbf{0.4726} & \textbf{0.5927} & 1.75 & \textbf{358.020} \\  %
    RW-\internalmodel & 0.8001 & 0.8851 & 0.9836 & 0.3958 & 0.4123 & \textbf{0.5961} & 1.10 & 669.747  \\  %
    \midrule
    \textbf{PA-\opensora-base} (ours) & 0.7656 & 0.8880 & 0.9859 & 0.5625 & 0.4582 & 0.5033 & 2.04 & 548.117 \\
    RN-\opensora-base & 0.7406 & 0.8820 & 0.9873 & 0.5750 & 0.4034 & 0.4464 & 5.19 & 600.690 \\
    \midrule
    StreamingSVD & \textbf{0.8172} & 0.8916 & \textbf{0.9929} & 0.65 & 0.4264 & 0.5566 & \textbf{1.08} & 440.272 \\
    \midrule
    SVD-XT & 0.6102 & 0.8136 & 0.9724 & \textbf{0.9875} & 0.3019 & 0.4814 & 2.10 & 702.343 \\
    \midrule
    FIFO-OSP & 0.7577 & 0.8990 & 0.9731 & 0.75 & N/A & 0.5675 & 18.32 & 975.459 \\
    \bottomrule
\end{tabular}

\end{table*}

\subsection{Implementation}

\noindent\textbf{Our models and baseline models}
We implement our progressive autoregressive video diffusion models by fine-tuning from pre-trained models. Specifically, we use two video diffusion models based on the diffusion transformer architecture~\cite{peebles2023dit,openai2024sora}: Open-Sora v1.2~\cite{opensora} (denoted as \opensora) and a modified variant of Open-Sora (denoted as \internalmodel in later experiments).
Both models are latent video diffusion models~\cite{blattmann2023align}, each utilizing a corresponding 3D VAE that encodes 17 (\opensora) or 16 (\internalmodel) raw video frames into 5 latent frames.
\opensora generates videos at 240$\times$424 resolution 24 FPS with 30 sampling steps.
\internalmodel produces results at 176$\times$320 resolution 24 FPS with 50 sampling steps.
Based on \opensora and \internalmodel, we also implement two baseline autoregressive video generation methods, \textit{replacement-with-noise} (denoted as RW) and \textit{replacement-without-noise} (denoted as RN) (\cref{sec:bg:long_replacement}), to compare with our proposed \textit{progressive autoregressive} (denoted as PA) video generation method (\cref{sec:method}).

We train \internalmodel on our progressive noise levels, as discussed in~\cref{sec:method:training}.
The resulting model can perform progressive autoregressive video generation, which we denote as PA-\internalmodel.
We also train \internalmodel with the \textit{replacement-with-noise} method, which we will denote as RW-\internalmodel.
Starting from the same pre-trained weight of the base model, RW-\internalmodel is trained for 3 times more training steps compared to PA-\internalmodel.

\opensora undergoes masked pre-training~\cite{opensora}, where the masked latent frames $\displaystyle \rvx_{0:E-1}^0$ are clean without any added noise~\cite{opensora}. 
This allows the \opensora base model to perform autoregressive video generation with the \textit{replacement-without-noise} method.
We denote this model as RN-\opensora-base.
Such training also allows \opensora to learn that the noise levels $\displaystyle \rvt_{0:F-1}$ can be independent with respect to the latent frames and thus enables our \textit{progressive autoregressive} video denoising sampling procedure (\cref{alg:main}) to work training-free.
We denote this model as PA-\opensora-base.
Please refer to~\cref{sec:training_details} for training details.

\subsection{Long video generation}
\label{sec:exp:long}
The baseline methods are described in~\cref{sec:appx:baselines}.

\noindent\textbf{Metrics}
We consider 6 metrics in VBench~\cite{huang2024vbench}: subject consistency, background consistency, motion smoothness, dynamic degree, aesthetic quality, and imaging quality. 
We compute average metrics using VBench-long, where each metric is computed on 30 2-second clips for each 60-second video; for subject and background consistency, a clip-to-clip metric is considered in addition to the average metric over the clips.
We also show how the metrics vary over time by plotting the metrics over the 30 2-second clips averaged over the 80 60-second videos.

Similar to~\cite{henschel2024streamingt2vconsistentdynamicextendable}, we also use the Adaptive Detector algorithm from PySceneDetect~\cite{pyscenedetect} to count the number of detected scene changes, where Num Scenes $=1$ means that there is no scene change detected.

We also compute Fréchet Video Distance (FVD)~\cite{unterthiner2018fvd} to measure the overall quality of the generated videos compared to real videos.
We adopt the improved implementation of FVD proposed in~\cite{ge2024cdfvd} using the VideoMAE-v2~\cite{wang2023videomaev2} model.
The FVD metric usually requires a large number of video samples in order to produce a reliable value.
Since our testing set includes only 40 real videos and each model only generate 80 videos, naively computing FVD on them results in erroneous values such as -3.62e+64.
Instead, we compute FVD on the 2-second clips of the long videos, so that we have 1495 real videos and 2400 generated videos.

\noindent\textbf{Quantitative Results}
We present the average metrics for each model in~\cref{tab:main}. 
The metrics are averaged over all the videos that each model generates from our testing set described above.
Our PA-\internalmodel has the best results overall. 
Notably, it surpasses other methods in FVD by a substantial margin, illustrating that its results are the most realistic.
It also achieves either the best or close-to-best in other metrics.
Its \textit{replacement-with-noise} counterpart, RW-\internalmodel, suffers from poor Dynamic Degree and FVD, because its videos are mostly static.
Our RW-\opensora-base surpasses its \textit{replacement-without-noise} counterpart RN-\opensora-base in all metrics except for being close at Dynamic Degree, while using the exact same model parameters without any finetuning.
RN-\opensora-base mainly suffers from a high number of scene changes.

\begin{figure}[htbp]
    \centering
    \includegraphics[width=0.99\linewidth]{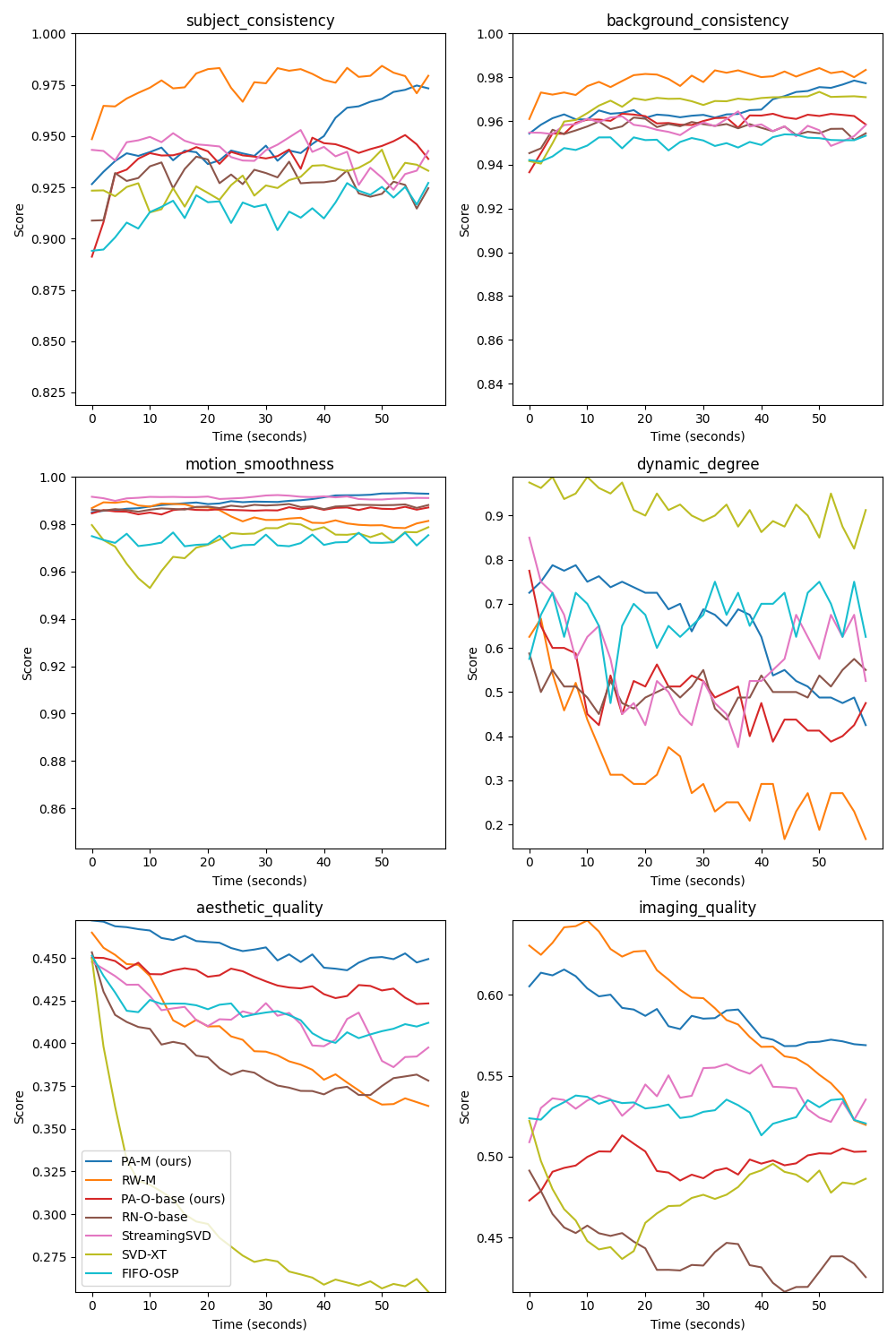}
    \caption{
    VBench~\citep{huang2024vbench} scores over the 60-second duration, which are computed on 30 2-second clips.
    }
    \label{fig:vbench}

\end{figure}

In~\cref{fig:vbench}, we also illustrate the trend of metrics over the 1-minute duration of videos for each model. 
Our models \internalmodel-PA and \opensora-PA can best maintain the level of all metrics, while their \textit{replacement}-method counterparts, \internalmodel-RW and \opensora-RN, both exhibit distinct reduction in dynamic degree, aesthetic quality, and imaging quality.

\begin{figure}[tbp]
    \centering
    \includegraphics[width=0.99\linewidth]{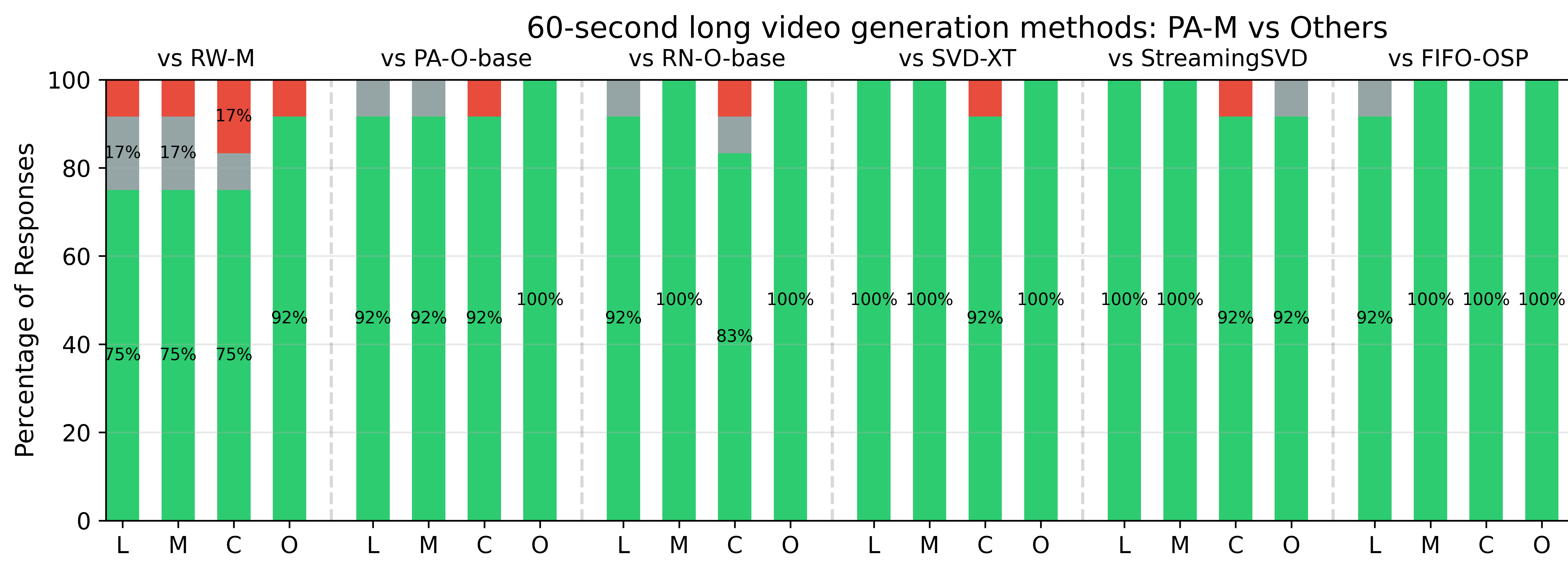}
    \vspace{-2mm}
    \caption{Human evaluation results comparing long video methods on long-shot (L), motion (M), temporal consistency (C), and overall (O).
    }
    \label{fig:user_study_main}
\end{figure}

\begin{figure*}[htbp]
    \centering
    
    \begin{minipage}{0.02\textwidth} %
        \rotatebox{90}{\small{\textbf{PA-\internalmodel}}} %
    \end{minipage}
    \begin{minipage}{0.161\textwidth}
        \includegraphics[width=\linewidth]{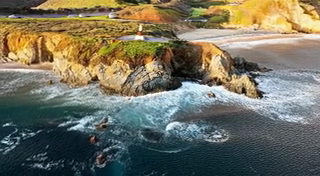}
    \end{minipage}\hfill
    \begin{minipage}{0.161\textwidth}
        \includegraphics[width=\linewidth]{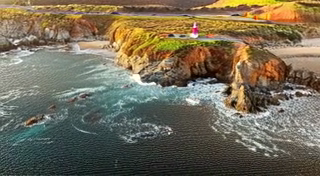}
    \end{minipage}\hfill
    \begin{minipage}{0.161\textwidth}
        \includegraphics[width=\linewidth]{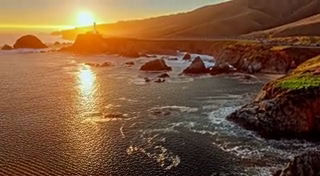}
    \end{minipage}\hfill
    \begin{minipage}{0.161\textwidth}
        \includegraphics[width=\linewidth]{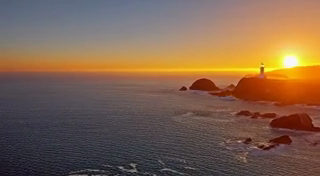}
    \end{minipage}\hfill
    \begin{minipage}{0.161\textwidth}
        \includegraphics[width=\linewidth]{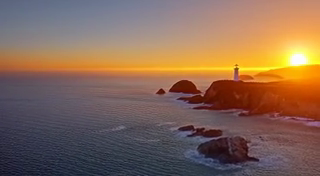}
    \end{minipage}\hfill
    \begin{minipage}{0.161\textwidth}
        \includegraphics[width=\linewidth]{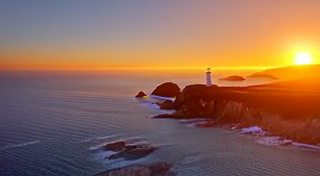}
    \end{minipage}\hfill

    \begin{minipage}{0.02\textwidth} %
        \rotatebox{90}{\small{RW-\internalmodel}} %
    \end{minipage}
    \begin{minipage}{0.161\textwidth}
        \includegraphics[width=\linewidth]{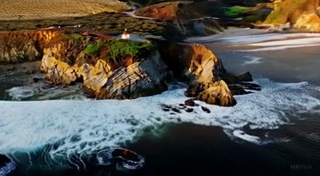}
    \end{minipage}\hfill
    \begin{minipage}{0.161\textwidth}
        \includegraphics[width=\linewidth]{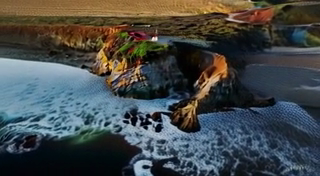}
    \end{minipage}\hfill
    \begin{minipage}{0.161\textwidth}
        \includegraphics[width=\linewidth]{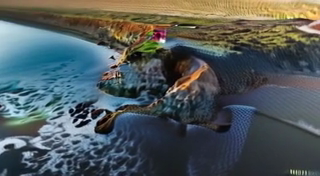}
    \end{minipage}\hfill
    \begin{minipage}{0.161\textwidth}
        \includegraphics[width=\linewidth]{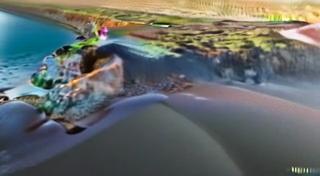}
    \end{minipage}\hfill
    \begin{minipage}{0.161\textwidth}
        \includegraphics[width=\linewidth]{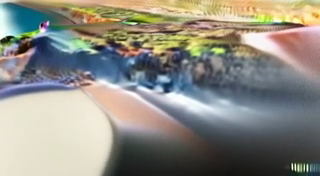}
    \end{minipage}\hfill
    \begin{minipage}{0.161\textwidth}
        \includegraphics[width=\linewidth]{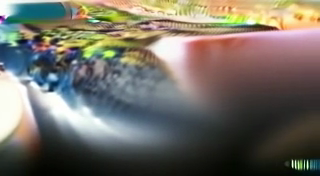}
    \end{minipage}\hfill
    
    \begin{minipage}{0.02\textwidth} %
        \rotatebox{90}{\small{\textbf{PA-\opensora-b}}} %
    \end{minipage}
    \begin{minipage}{0.161\textwidth}
        \includegraphics[width=\linewidth]{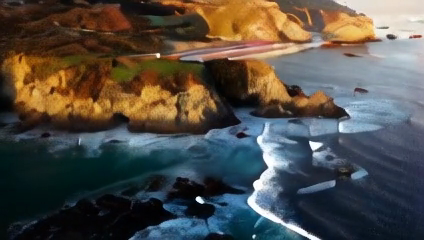}
    \end{minipage}\hfill
    \begin{minipage}{0.161\textwidth}
        \includegraphics[width=\linewidth]{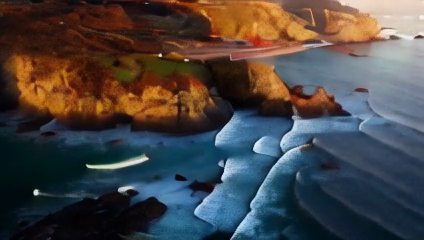}
    \end{minipage}\hfill
    \begin{minipage}{0.161\textwidth}
        \includegraphics[width=\linewidth]{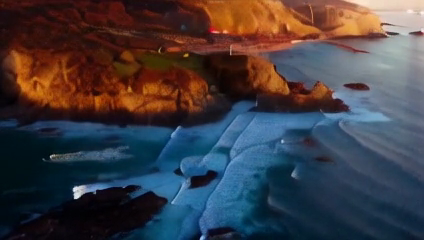}
    \end{minipage}\hfill
    \begin{minipage}{0.161\textwidth}
        \includegraphics[width=\linewidth]{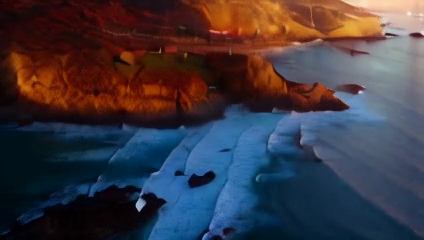}
    \end{minipage}\hfill
    \begin{minipage}{0.161\textwidth}
        \includegraphics[width=\linewidth]{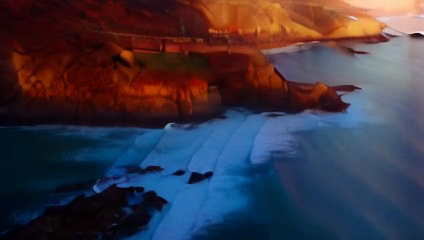}
    \end{minipage}\hfill
    \begin{minipage}{0.161\textwidth}
        \includegraphics[width=\linewidth]{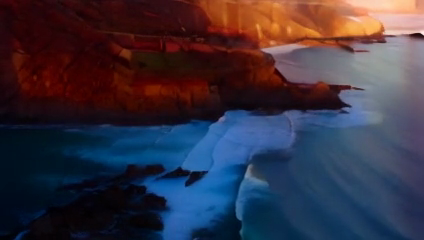}
    \end{minipage}\hfill

    \begin{minipage}{0.02\textwidth} %
        \rotatebox{90}{\small{RN-\opensora-b}} %
    \end{minipage}
    \begin{minipage}{0.161\textwidth}
        \includegraphics[width=\linewidth]{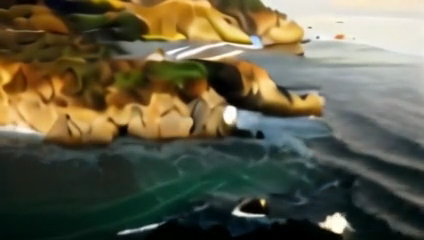}
    \end{minipage}\hfill
    \begin{minipage}{0.161\textwidth}
        \includegraphics[width=\linewidth]{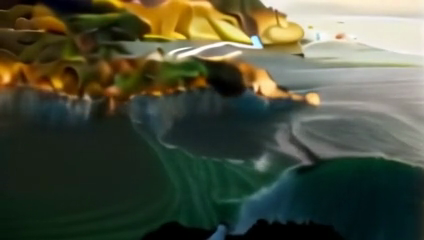}
    \end{minipage}\hfill
    \begin{minipage}{0.161\textwidth}
        \includegraphics[width=\linewidth]{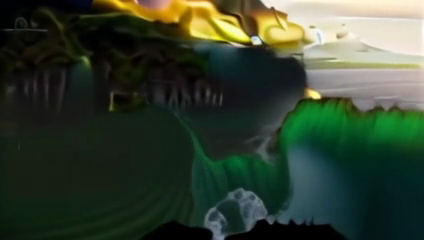}
    \end{minipage}\hfill
    \begin{minipage}{0.161\textwidth}
        \includegraphics[width=\linewidth]{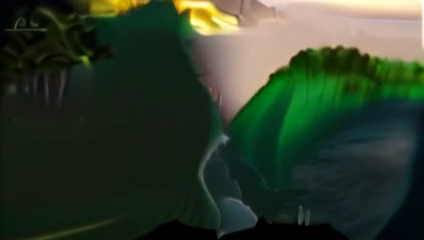}
    \end{minipage}\hfill
    \begin{minipage}{0.161\textwidth}
        \includegraphics[width=\linewidth]{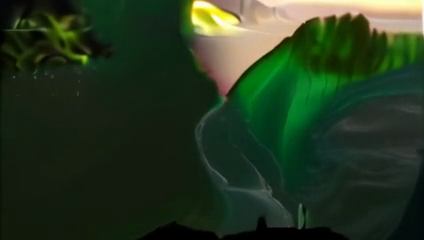}
    \end{minipage}\hfill
    \begin{minipage}{0.161\textwidth}
        \includegraphics[width=\linewidth]{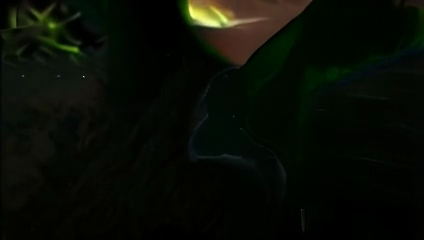}
    \end{minipage}\hfill

    \begin{minipage}{0.02\textwidth} %
        \rotatebox{90}{\small{S-T2V}} %
    \end{minipage}
    \begin{minipage}{0.161\textwidth}
        \includegraphics[width=\linewidth]{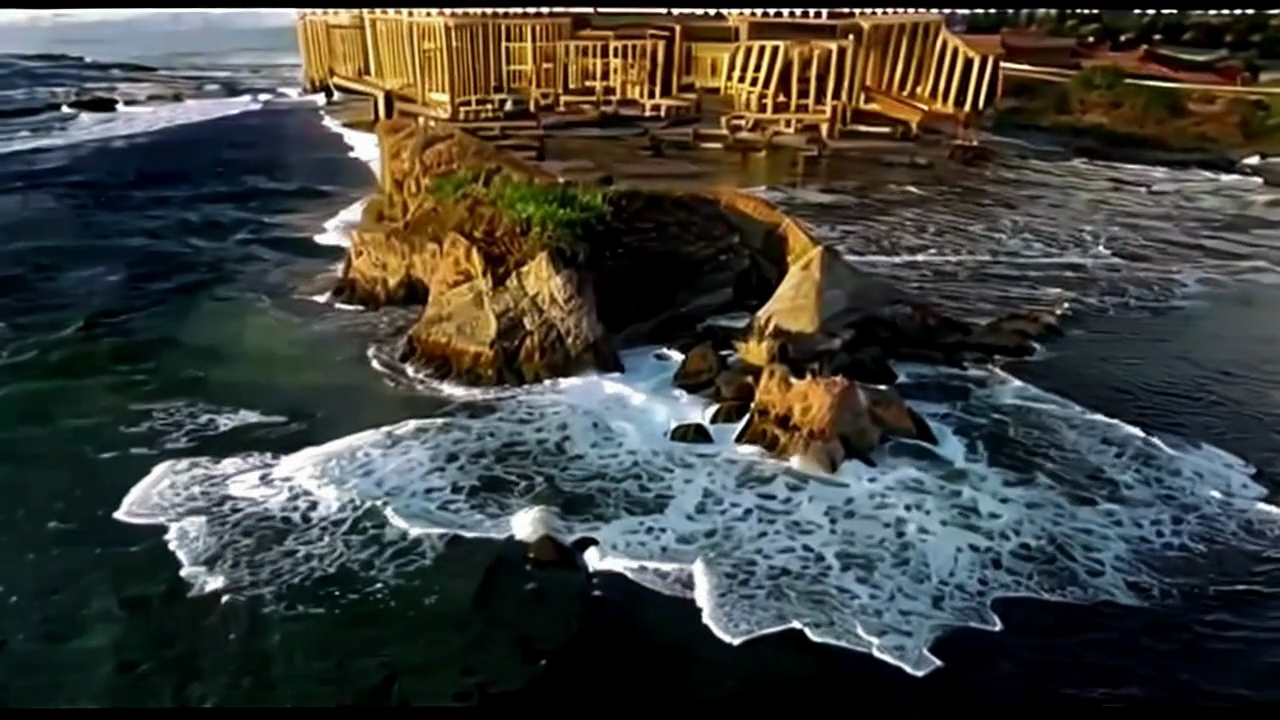}
    \end{minipage}\hfill
    \begin{minipage}{0.161\textwidth}
        \includegraphics[width=\linewidth]{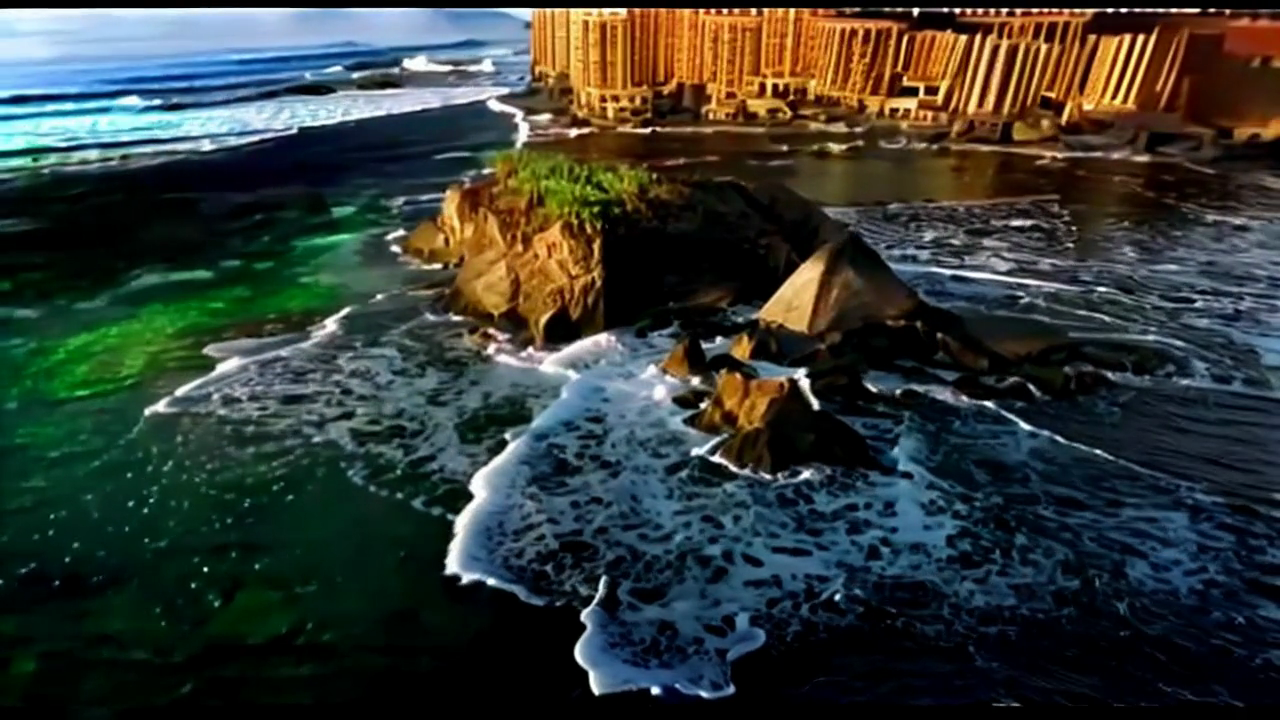}
    \end{minipage}\hfill
    \begin{minipage}{0.161\textwidth}
        \includegraphics[width=\linewidth]{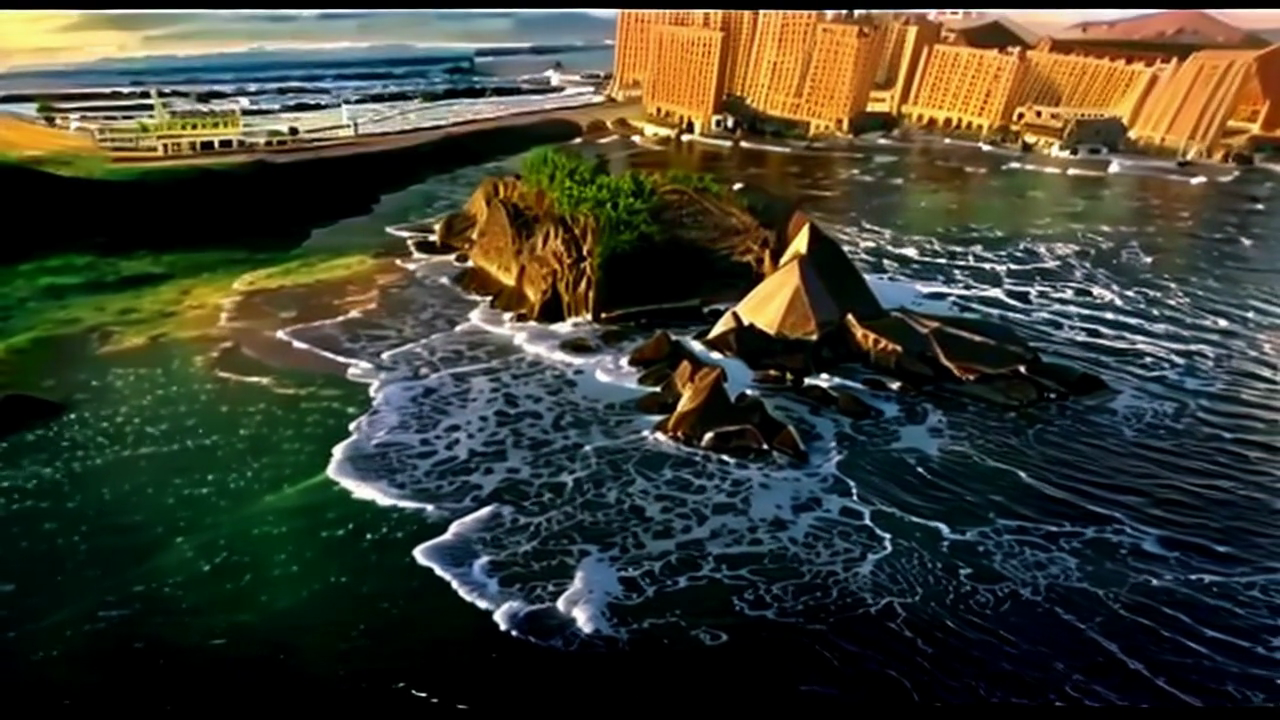}
    \end{minipage}\hfill
    \begin{minipage}{0.161\textwidth}
        \includegraphics[width=\linewidth]{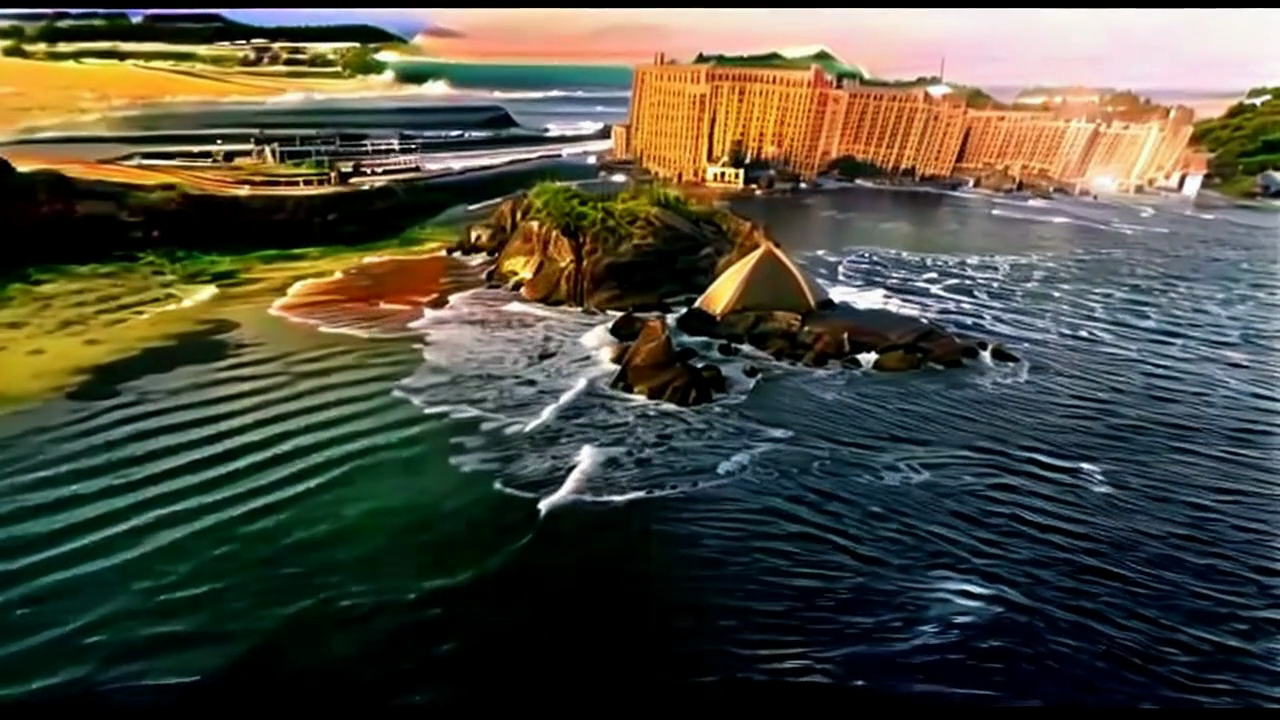}
    \end{minipage}\hfill
    \begin{minipage}{0.161\textwidth}
        \includegraphics[width=\linewidth]{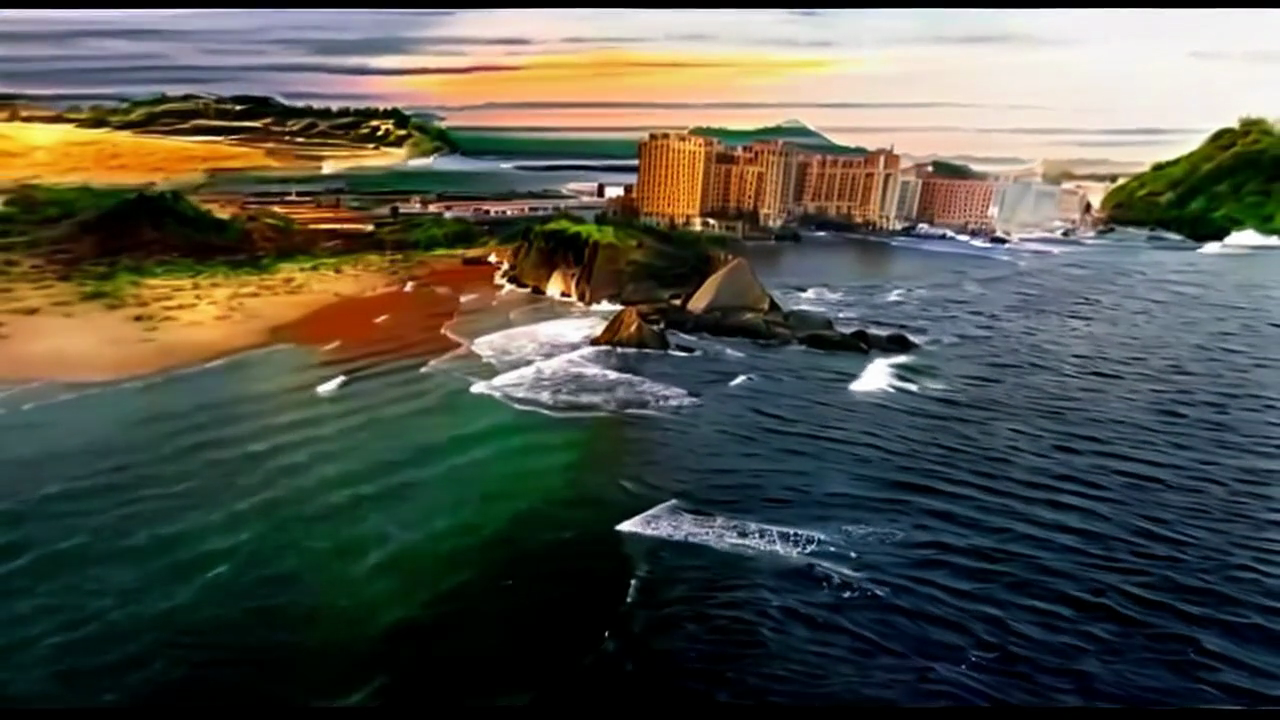}
    \end{minipage}\hfill
    \begin{minipage}{0.161\textwidth}
        \includegraphics[width=\linewidth]{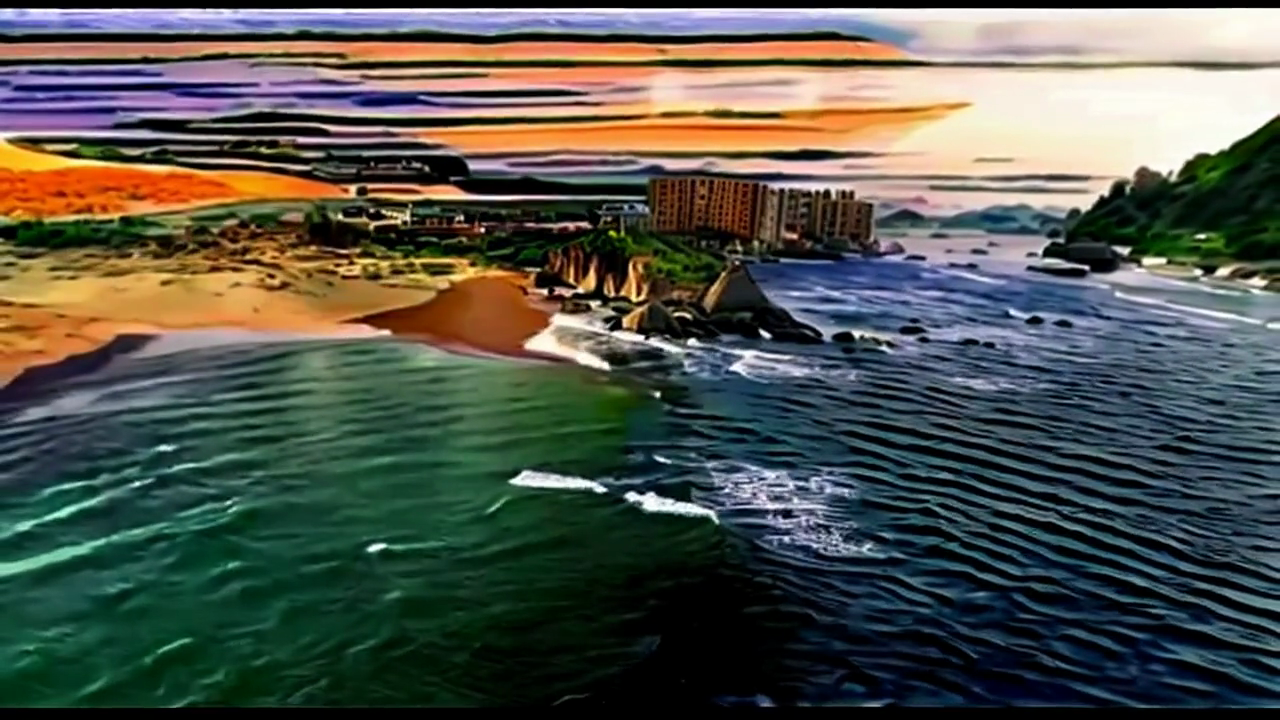}
    \end{minipage}\hfill

    \begin{minipage}{0.02\textwidth} %
        \rotatebox{90}{\small{SVD}} %
    \end{minipage}
    \begin{minipage}{0.161\textwidth}
        \includegraphics[width=\linewidth]{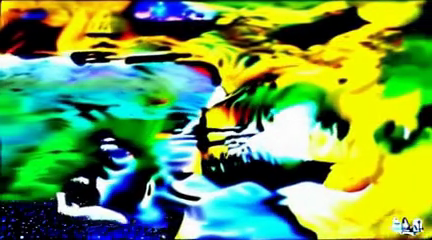}
    \end{minipage}\hfill
    \begin{minipage}{0.161\textwidth}
        \includegraphics[width=\linewidth]{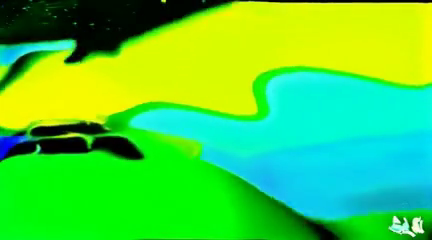}
    \end{minipage}\hfill
    \begin{minipage}{0.161\textwidth}
        \includegraphics[width=\linewidth]{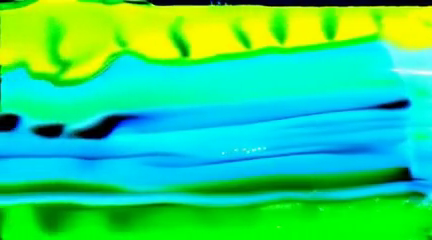}
    \end{minipage}\hfill
    \begin{minipage}{0.161\textwidth}
        \includegraphics[width=\linewidth]{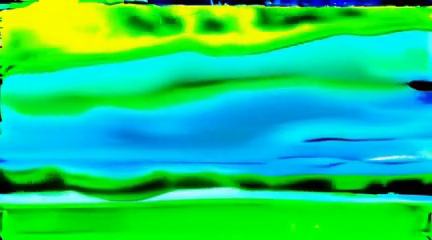}
    \end{minipage}\hfill
    \begin{minipage}{0.161\textwidth}
        \includegraphics[width=\linewidth]{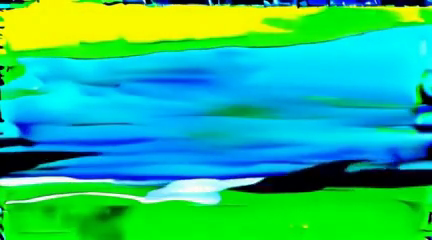}
    \end{minipage}\hfill
    \begin{minipage}{0.161\textwidth}
        \includegraphics[width=\linewidth]{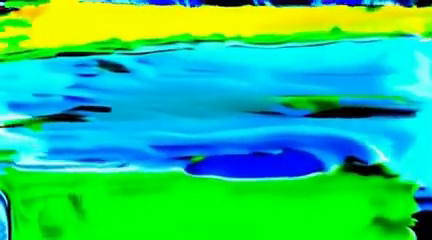}
    \end{minipage}\hfill

    \begin{minipage}{0.02\textwidth} %
        \rotatebox{90}{\small{FIFO}} %
    \end{minipage}
    \begin{minipage}{0.161\textwidth}
        \includegraphics[width=\linewidth]{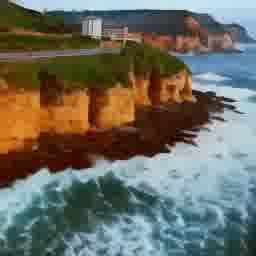}
    \end{minipage}\hfill
    \begin{minipage}{0.161\textwidth}
        \includegraphics[width=\linewidth]{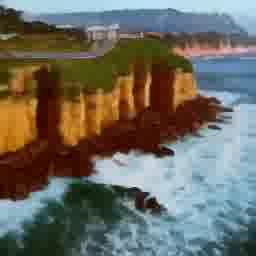}
    \end{minipage}\hfill
    \begin{minipage}{0.161\textwidth}
        \includegraphics[width=\linewidth]{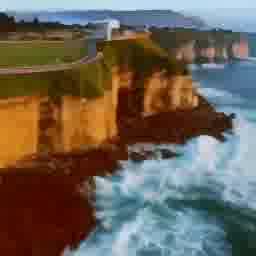}
    \end{minipage}\hfill
    \begin{minipage}{0.161\textwidth}
        \includegraphics[width=\linewidth]{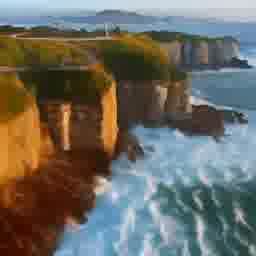}
    \end{minipage}\hfill
    \begin{minipage}{0.161\textwidth}
        \includegraphics[width=\linewidth]{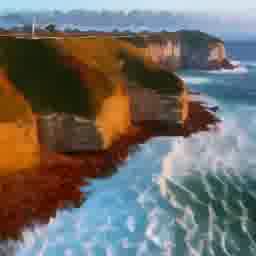}
    \end{minipage}\hfill
    \begin{minipage}{0.161\textwidth}
        \includegraphics[width=\linewidth]{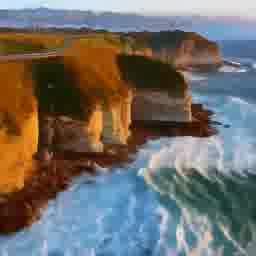}
    \end{minipage}\hfill

    \caption{
    Qualitative comparison of
    PA-\internalmodel (ours), RW-\internalmodel, PA-\opensora-base (ours), RN-\opensora-base, StreamingSVD from StreamingT2V~\cite{henschel2024streamingt2vconsistentdynamicextendable}, SVD-XT from Stable Video Diffusion~\cite{blattmann2023stable}, and FIFO-Diffusion~\cite{kim2024fifo}.
    Frames are evenly sampled from 1 minute long generated video, i.e. at 10, 20, 30, 40, 50, and 60 seconds.
    Our models can autoregressively generate 60-second, 1440-frame videos without quality degradation.
    }
    \label{fig:qualitative}
\end{figure*}

\begin{figure*}[htbp]
    \centering
    
    \begin{minipage}{0.02\textwidth} %
        \rotatebox{90}{\small{Full}} %
    \end{minipage}
    \begin{minipage}{0.161\textwidth}
        \includegraphics[width=\linewidth]{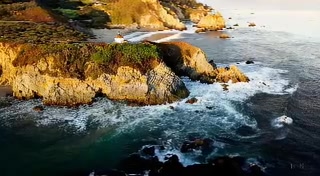}
    \end{minipage}\hfill
    \begin{minipage}{0.161\textwidth}
        \includegraphics[width=\linewidth]{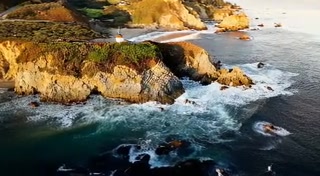}
    \end{minipage}\hfill
    \begin{minipage}{0.161\textwidth}
        \includegraphics[width=\linewidth]{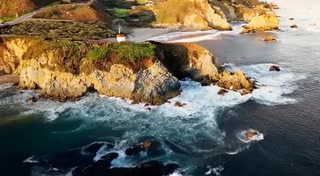}
    \end{minipage}\hfill
    \begin{minipage}{0.161\textwidth}
        \includegraphics[width=\linewidth]{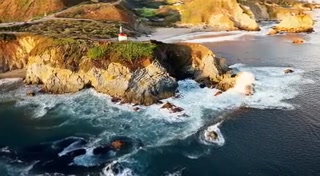}
    \end{minipage}\hfill
    \begin{minipage}{0.161\textwidth}
        \includegraphics[width=\linewidth]{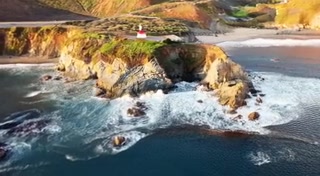}
    \end{minipage}\hfill
    \begin{minipage}{0.161\textwidth}
        \includegraphics[width=\linewidth]{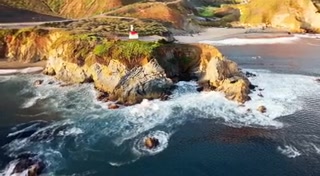}
    \end{minipage}\hfill

    \vspace{1em}

    \begin{minipage}{0.02\textwidth} %
        \rotatebox{90}{\small{Ablation 1}} %
    \end{minipage}
    \begin{minipage}{0.161\textwidth}
        \includegraphics[width=\linewidth]{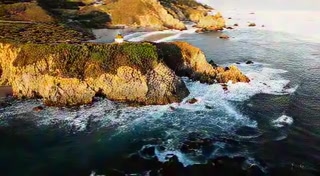}
    \end{minipage}\hfill
    \begin{minipage}{0.161\textwidth}
        \includegraphics[width=\linewidth]{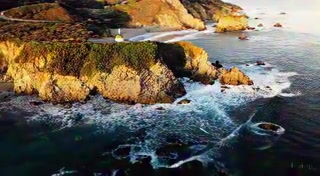}
    \end{minipage}\hfill
    \begin{minipage}{0.161\textwidth}
        \includegraphics[width=\linewidth]{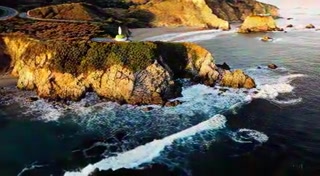}
    \end{minipage}\hfill
    \begin{minipage}{0.161\textwidth}
        \includegraphics[width=\linewidth]{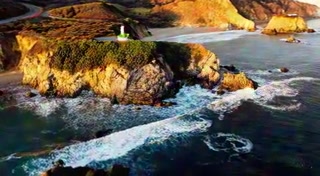}
    \end{minipage}\hfill
    \begin{minipage}{0.161\textwidth}
        \includegraphics[width=\linewidth]{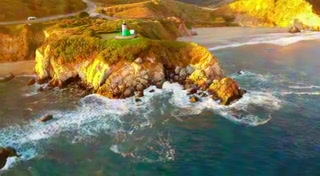}
    \end{minipage}\hfill
    \begin{minipage}{0.161\textwidth}
        \includegraphics[width=\linewidth]{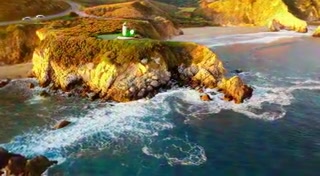}
    \end{minipage}\hfill

    \vspace{1em}

    \begin{minipage}{0.02\textwidth} %
        \rotatebox{90}{\small{Ablation 2}} %
    \end{minipage}
    \begin{minipage}{0.161\textwidth}
        \includegraphics[width=\linewidth]{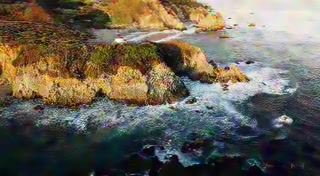}
    \end{minipage}\hfill
    \begin{minipage}{0.161\textwidth}
        \includegraphics[width=\linewidth]{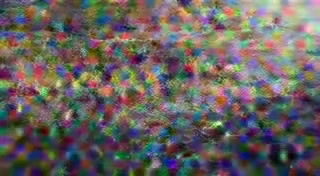}
    \end{minipage}\hfill
    \begin{minipage}{0.161\textwidth}
        \includegraphics[width=\linewidth]{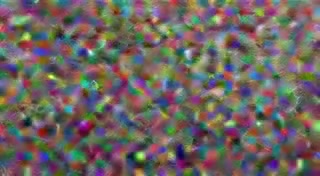}
    \end{minipage}\hfill
    \begin{minipage}{0.161\textwidth}
        \includegraphics[width=\linewidth]{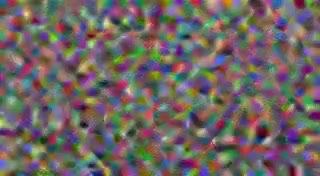}
    \end{minipage}\hfill
    \begin{minipage}{0.161\textwidth}
        \includegraphics[width=\linewidth]{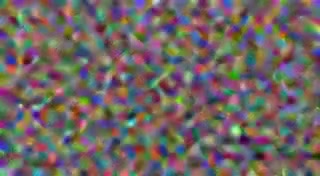}
    \end{minipage}\hfill
    \begin{minipage}{0.161\textwidth}
        \includegraphics[width=\linewidth]{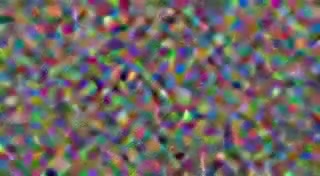}
    \end{minipage}\hfill
    
    \caption{
    Qualitative comparison for ablation study. Full represents for our full solution based on PA-\internalmodel, Ablation 1 is with \textit{chunked frames} but without \textit{overlapped conditioning}. Ablation 2 is without both techniques.
    The frames are evenly sampled from 16-second generated videos.
    }
    \label{fig:vs_ablation}
\end{figure*}

\noindent\textbf{Qualitative Results}
We also show strength of our method with qualitative comparison results in~\cref{fig:qualitative}. 
Both of our models demonstrate strong performance in terms of frame fidelity and motion realism (e.g. camera motion, wave motion, and running gestures) and outperforms other baselines. 
For more qualitative results, please refer to our project webpage.

\noindent\textbf{User study}
We conduct a human evaluation with 12 users to compare the generated videos from each method.
As shown in~\cref{fig:user_study_main}, our PA-\internalmodel is favored in each duel by a large margin.

\subsection{Ablation Study}
\label{sec:exp:ablation}

We conduct ablation studies on the PA-\internalmodel model to evaluate the impact of chunked frames (\cref{sec:method:chunked_latents}), and overlapped conditioning (\cref{sec:method:overlapped_conditioning}). 
Qualitative comparison is shown in~\cref{fig:vs_ablation} and in the project webpage. 
In Ablation 1, we observe that the absence of clean frames in the input sequence prevents noisy frames from attending to previous clean frames, resulting in poor performance over a long duration. 
This also causes frame-to-frame discontinuity, which is more noticeable in the project webpage. In Ablation 2, not decoding the video chunk-by-chunk leads to severe cumulative errors, causing the video to diverge after only a few seconds.

See~\cref{sec:add_ablation} for additional ablation study on variable length and the number of sampling steps $S$.

\section{Conclusion}
In this work, we target long video generation, a fundamental challenge of current video diffusion models.
We show that they can be naturally adapted to become progressive autoregressive video diffusion models without changing the architectures.
With our progressive noise levels and the autoregressive video denoising process (\cref{sec:method:progressive}), we achieve state-of-the-art results on 60-second long video generation.
Since our method does not require model architecture changes, it can be seamlessly combined with orthogonal works, paving the way for generating longer videos at higher quality, long-term dependency, and controllability.

\section{Acknowledgments}
This research was supported in part by NSF award IIS2107224 and ONR award N000142312124.

{
    \clearpage
    \small
    \bibliographystyle{cvpr2025/ieeenat_fullname}
    \bibliography{main_bib}
}

\clearpage
\appendix
\maketitlesupplementary

\section{Summary}
In this Appendix, we cover parallel works in~\cref{sec:concurrent}, related works in~\cref{sec:related}, limitations and discussions in~\cref{sec:limit_discuss}, training details in~\cref{sec:training_details}, evaluation details in~\cref{sec:eval_details}, additional qualitative results in~\cref{sec:supple:qualitative}, and additional ablation study in~\cref{sec:add_ablation}.

\section{Parallel Works}
\label{sec:concurrent}
The core idea of our PA-VDM is to 1. \textit{assign progressively increasing noise levels to the $F$ frames in the attention window} and 2. \textit{autoregressively apply the video diffusion model on progressively noised frames to generate long videos}.
The first part is inspired by Diffusion Forcing~\cite{chen2024diffusionforcingnexttokenprediction}, which proposes to assign independent per-frame noise levels to some frames rather than a single noise level.
We began developing our work right after July 1st, 2024, when Diffusion Forcing~\cite{chen2024diffusionforcingnexttokenprediction} was released on arXiv.
The first version of our preprint was submitted to arXiv on October 10th, 2024.
During this period, our work was developed independently, without the knowledge of two papers, Rolling Diffusion~\cite{ruhe2024rolling} and FIFO-Diffusion~\cite{kim2024fifo}.
While Rolling Diffusion, FIFO-Diffusion, and PA-VDM arrive at a similar high-level idea in parallel, the three methods have different focuses, naming and framing of the idea, implementation details, experimental setups, and final result quality.

Compared to~\cite{ruhe2024rolling,fan2024reinforcement}, PA-VDM:
\begin{enumerate}
    \item shows that it is possible to adapt a pre-trained video diffusion model to the progressive noise level schedule through finetuning, thus avoiding the otherwise immensely expensive computation cost of pre-training video diffusion models.~\cite{ruhe2024rolling} is trained from scratch and~\cite{kim2024fifo} is training-free.
    \item achieves state-of-the-art 60-second long video generation at a quality comparable to frontier video diffusion models, demonstrating much longer video length and better quality than~\cite{ruhe2024rolling,kim2024fifo}.
\end{enumerate}
We provide comparisons between our models (PA-\internalmodel and PA-\opensora) and~\cite{kim2024fifo} on our 60-second long video generation benchmark (\cref{sec:testing_set}) in~\cref{sec:exp:long,tab:main}. 
Our method achieves substantially better qualitative and quantitative results than~\cite{kim2024fifo}. 
\cite{kim2024fifo} also requires doubled inference cost, while our method only requires additional inference cost that is a fraction of the original cost (10\% for PA-\internalmodel and 16.66\% for PA-\opensora).
We do not compare PA-VDM with~\cite{ruhe2024rolling} as there is no released code and it does not support text-conditioned open-domain generation.

\section{Related Works}
\label{sec:related}
The field of long video generation has faced significant challenges due to the computational complexity and resource constraints associated with training models on longer videos. As a result, most existing text-to-video diffusion models \cite{guo2023animatediff,ho2022imagen,ho2022videodiffusionmodels,blattmann2023stable} have been limited to generating fixed-size video clips, which leads to noticeable degradation in quality when attempting to generate longer videos. Recent works are proposed to address these challenges through innovative approaches that either extend existing models or introduce novel architectures and fusion methods. 

Freenoise~\cite{qiu2024freenoisetuningfreelongervideo} utilizes sliding window temporal attention to ensure smooth transitions between video clips but falls short in maintaining global consistency across long video sequences. Gen-L-video~\cite{wang2023genlvideomultitextlongvideo}, on the other hand, decomposes long videos into multiple short segments, decodes them in parallel using short video generation models, and later applies an optimization step to align the overlapping regions for continuity. FreeLong~\cite{lu2024freelong} introduces a sophisticated approach which balances the frequency distribution of long video features in different frequency during the denoising process. Vid-GPT~\cite{gao2024vidgptintroducinggptstyleautoregressive} introduces GPT-style autoregressive causal generation for long videos.

More recently, Short-to-Long (S2L) approaches are proposed, where correlated short videos are firstly generated and then smoothly transit in-between to form coherent long videos. StreamingT2V~\cite{henschel2024streamingt2vconsistentdynamicextendable} adopts this strategy by introducing the conditional attention and appearance preservation modules to capture content information from previous frames, ensuring consistency with the starting frames. It further enhances the visual coherence by blending shared noisy frames in overlapping regions, similar to the approach used by SEINE~\cite{chen2023seine}. NUWA-XL~\cite{yin2023nuwa} leverages a hierarchical diffusion model to generate long videos using a coarse-to-fine approach, progressing from sparse key frames to denser intermediate frames. However, it has only been evaluated on a cartoon video dataset rather than natural videos. VideoTetris~\cite{tian2024videotetris} introduces decomposing prompts temporally and leveraging a spatio-temporal composing module for compositional video generation. 

Another line of research focuses on controllable video generation~\cite{zhuang2024vlogger,tian2024emo,hu2024animate,zhu2024champ} and has proposed solutions for long video generation using overlapped window frames. These approaches condition diffusion models using both frames from previous windows and signals from the current window. While these methods demonstrate promising results in maintaining consistent appearances and motions, they are limited to their specific application domains which relies heavily on strong conditional inputs.

\todos{MAGViT, Long Video GAN}

\section{Limitations and discussions}
\label{sec:limit_discuss}
A limitation of our method is the demand of a well-trained base video diffusion model.
Similar to the \textit{replacement} methods~\cite{ho2022videodiffusionmodels,opensora} and other approaches like StreamingT2V~\cite{henschel2024streamingt2vconsistentdynamicextendable}, our method autoregressively applies a video diffusion model to generate long videos. 
Such autoregressive video generation poses huge challenge on the base video diffusion model. 
Some slight errors remaining in the ``clean'' frames $\displaystyle \rvx^0$ may not be noticeable in a single video clip; 
however, in the autoregressive scenario, these error can be carried onto later frames, resulting in quality degradation.
Further more, as the video diffusion model is only trained on denoising latent frames of real video data, it may poorly handle such distribution shift towards the generated erroneous frames~\cite{xie2024carve3d,fan2024reinforcement}, resulting in more severe quality drop.
This means that even after finetuning on our progressive noise levels, our method could still generate videos with some degree of quality degradation close to the ending, if the base video diffusion model is not well trained.
Among the qualitative videos generated by our PA-\internalmodel, in some cases, the video quality slightly degrades in the last 10 seconds.

Another limitation of our method is the subtle temporal flickering happening about every second in our PA-\internalmodel results.
It is caused by a flaw in the backbone video diffusion model \internalmodel's 3D VAE, as evident by the presence of such flickering in both PA-\internalmodel and RW-\internalmodel results while no such flickering is present in the PA-\opensora results.

There are many promising future directions to extend this work.
We only train on progressively increasing noise levels to reduce the space of noise levels for easier convergence.
If sufficient computing resources are available, training on fully random, per-frame independent noise levels would enable a single model for various tasks with arbitrary lengths, including video extension, connection, temporal super-resolution.
Another promising future application of the long video generation ability of our models is to use them as world simulators, useful for tasks in robotics and 3D vision.
Being able to generate long videos without quality degradation is an substantial step towards this direction.
\todos{citations}

\section{Training details}
\label{sec:training_details}
\internalmodel is pre-trained on captioned image and video datasets, containing 1 million videos and 2.3 billion images.
These data are licensed and have been filtered to remove low-quality content.
We train PA-\internalmodel on video clips of $16, 32, ..., 176$ raw frames that correspond to $F=5, 10, ..., 55$ latent frames.
The $F=55$ attention window length is derived by setting $F=S+5$, where $S=50$ is the number of sampling steps in M ($S=30$ in O) and $5$ is the length of an additional chunk of latent frames, as described in~\cref{sec:method:chunked_latents,sec:method:overlapped_conditioning}. 
The shorter latent frame lengths $F=5, 10, ..., 50$ are used for the variable length training, as discussed in~\cref{sec:method:variable_length}.
RW-\internalmodel is trained on videos of $64$ frames that corresponds to $F=20$ frames.

\subsection{Modification to the base model}
To implement progressive autoregressive video diffusion models on top of their pre-trained foundation video diffusion models, we do not need to modify the base model architectures.
Instead, we only need to modify the model's forward, training, and inference procedures.
In the training and inference procedures, we replace the single noise level $\displaystyle \rt \in [0, T)$ from regular diffusion model training~\cite{ho2020ddpm,ho2022videodiffusionmodels} with our per-frame noise level $\displaystyle \rvt_{0:F-1}$ and $\displaystyle \bm{\tau}'_{0:S-1}$ (\cref{sec:method:training,sec:method:progressive}).
To accommodate this change, we only need to make a single modification to the the noise level embedding computation in the model's forward procedure. 
While the regular timestep only has the batch size dimension $B$, our progressive timesteps has two dimensions $B, F$.
We first flatten them into the batch dimension of size $B \times F$, pass it to the timesteps embedding module, unflatten the two dimensions, and finally broadcast the timestep embedding to the same shape of the frames so they can be combined through either addition, concatenation, modulation, or cross-attention~\cite{perez2018film,NIPS2017_attention,peebles2023dit}.

\section{Evaluation details}
\label{sec:eval_details}
\subsection{Baselines}
\label{sec:appx:baselines}
As discussed in~\cref{sec:4.1}, using our base models, we implement two baseline autoregressive video generation methods on three models, which are denoted as RW-\internalmodel, RN-\opensora-base, and RN-\opensora.
We also compare to Stable Video Diffusion (SVD)~\cite{blattmann2023stable} and StreamingT2V~\cite{henschel2024streamingt2vconsistentdynamicextendable} model families.
Specifically, we consider the SVD-XT model from SVD, a image-to-video model that generates a short video clip of 25 frames at 576x1024 resolution given an conditioning image.
We apply it autoregressively, using the last image of the previous clip as the condition for generating a new clip.
This is equivalent to the \textit{replacement-without-noise} method except that it only conditions on a single frame rather than a chunk of 17 frames as RN-\opensora.
We also consider the StreamingSVD model from StreamingT2V, a image-to-long-video generation model that uses SVD as the base model~\cite{henschel2024streamingt2vconsistentdynamicextendable}; its autoregressive video generation is enabled by training additional modules that connect to the base model via cross-attention.
Similar to our progressive autoregressive video diffusion models, StreamingSVD can autoregressively generate long videos at 720x1280 resolution with arbitrary lengths, which we set to 1440 frames.
We also compare to a concurrent work FIFO-Diffusion~\cite{kim2024fifo} implemented on Open-Sora-Plan v1.0.0~\cite{pku_open_sora_plan}, denoted as FIFO-OSP.
It generates at 256x256 resolution with a context window of 65 latent frames.
See~\cref{sec:concurrent} for a discussion on~\cite{kim2024fifo} and other concurrent works.
See~\cref{sec:eval_details} for details on our testing set, quantitative metrics, and traditional video quality evaluation.

\paragraph{FIFO-OSP}
FIFO-Diffusion~\cite{kim2024fifo} is a parallel work that adopts a similar high-level idea as our method on pre-trained video diffusion models without any fine-tuning (see more discussion in~\cref{sec:concurrent}).
It provides training-free implementations on VideoCrafter2 and Open-Sora-Plan v1.1.0~\cite{pku_open_sora_plan}. 
We choose its Open-Sora-Plan implementation since our method is also implemented on DiT-base~\cite{peebles2023dit} models, \internalmodel and Open-Sora (\opensora)~\cite{opensora}.
Open-Sora-Plan v1.1.0 generate videos at 512x512 resolution.
Since there is no distributed inference support in the released code of FIFO-Diffusion, we adopt Open-Sora-Plan v1.0.0 in our reproduced FIFO-Diffusion results in order to saving computation costs by inferencing at the 256x256 resolution instead of the original 512x512 resolution.

\subsection{Testing set}
\label{sec:testing_set}
\paragraph{Text prompts and real videos}
Our testing set consists of 40 text prompts and the corresponding real videos, sampled from Sora~\cite{opensora} demo videos, MiraData~\cite{ju2024miradatalargescalevideodataset}, 
UCF-101~\cite{soomro2012ucf101}, and LOVEU~\cite{wu2023tune,wu2023cvpr}.
For each text prompt, we generate two videos with 1440 frames, 60 seconds long at 24 FPS, resulting in a total of 80 videos.
We use these 80 videos from each model for both quantitative and qualitative results, unless specified otherwise.
Due to computation resource limitations of sampling 1-minute long videos, we only obtained partial results from \internalmodel-PA, StreamingSVD and FIFO-OSP, including 48, 40, 40 videos from 24, 40, 40 text prompts respectively. 
This testing set measures the zero-shot long video generation ability of the models, since none of them are specifically trained on any of the above datasets.

\paragraph{Real video initialization}
Since our focus is on long video generation, we focus on the video extension capability of the models rather than the text-to-short-video generation capability.
Thus, we use the initial frames of the videos as the condition for all models, similar to the setting in~\cite{henschel2024streamingt2vconsistentdynamicextendable}.
\internalmodel, \opensora~\cite{opensora}, StreamingSVD~\cite{henschel2024streamingt2vconsistentdynamicextendable}, SVD-XT~\cite{blattmann2023stable}, and FIFO-OSP~\cite{kim2024fifo,pku_open_sora_plan} use 16, 17, 1, 1, and 65 frames from the real video as the initial condition.
Note that our PA-\internalmodel and PA-\opensora only require one chunk of frames (16 and 17 for \internalmodel and \opensora respectively), which is substantially less than the full context window of 65 frames required by FIFO-Diffusion~\cite{kim2024fifo}.
This advantage is obtained from our variable-length autoregressive generation design as described in~\cref{sec:method:variable_length}.

\section{Additional Qualitative Results}
\label{sec:supple:qualitative}
\begin{figure*}[htbp]
    \centering
    
    \begin{minipage}{0.02\textwidth} %
        \rotatebox{90}{\small{\textbf{PA-\internalmodel}}} %
    \end{minipage}
    \begin{minipage}{0.161\textwidth}
        \includegraphics[width=\linewidth]{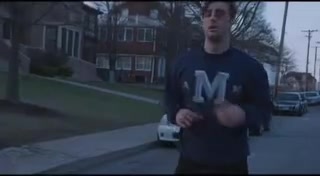}
    \end{minipage}\hfill
    \begin{minipage}{0.161\textwidth}
        \includegraphics[width=\linewidth]{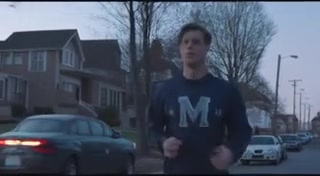}
    \end{minipage}\hfill
    \begin{minipage}{0.161\textwidth}
        \includegraphics[width=\linewidth]{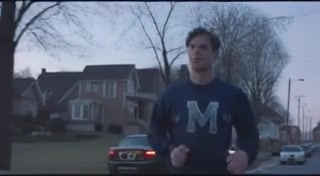}
    \end{minipage}\hfill
    \begin{minipage}{0.161\textwidth}
        \includegraphics[width=\linewidth]{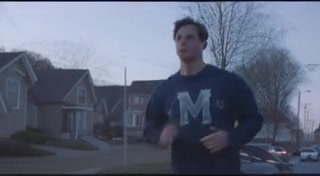}
    \end{minipage}\hfill
    \begin{minipage}{0.161\textwidth}
        \includegraphics[width=\linewidth]{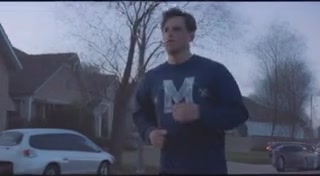}
    \end{minipage}\hfill
    \begin{minipage}{0.161\textwidth}
        \includegraphics[width=\linewidth]{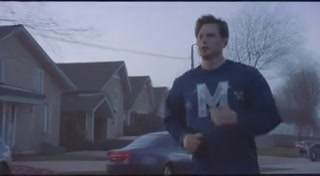}
    \end{minipage}\hfill
    
    \begin{minipage}{0.02\textwidth} %
        \rotatebox{90}{\small{\textbf{PA-\opensora-b}}} %
    \end{minipage}
    \begin{minipage}{0.161\textwidth}
        \includegraphics[width=\linewidth]{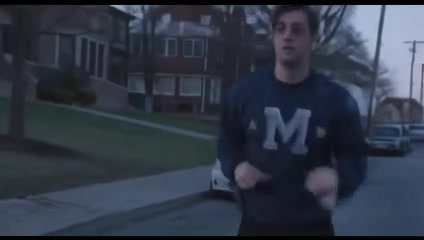}
    \end{minipage}\hfill
    \begin{minipage}{0.161\textwidth}
        \includegraphics[width=\linewidth]{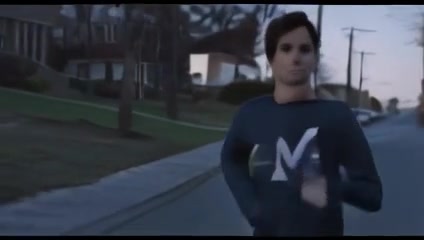}
    \end{minipage}\hfill
    \begin{minipage}{0.161\textwidth}
        \includegraphics[width=\linewidth]{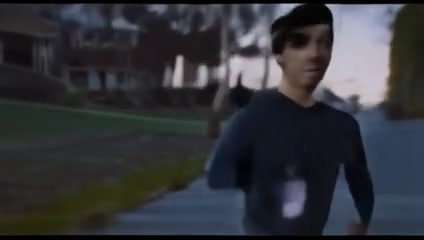}
    \end{minipage}\hfill
    \begin{minipage}{0.161\textwidth}
        \includegraphics[width=\linewidth]{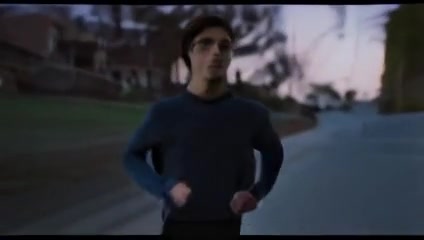}
    \end{minipage}\hfill
    \begin{minipage}{0.161\textwidth}
        \includegraphics[width=\linewidth]{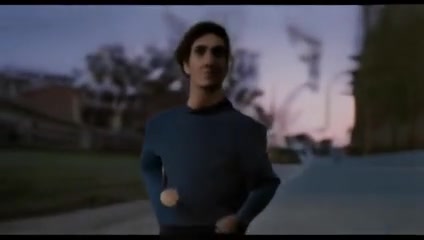}
    \end{minipage}\hfill
    \begin{minipage}{0.161\textwidth}
        \includegraphics[width=\linewidth]{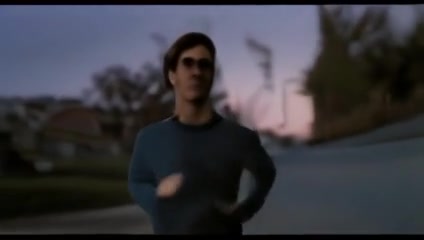}
    \end{minipage}\hfill

    \begin{minipage}{0.02\textwidth} %
        \rotatebox{90}{\small{RN-\opensora-b}} %
    \end{minipage}
    \begin{minipage}{0.161\textwidth}
        \includegraphics[width=\linewidth]{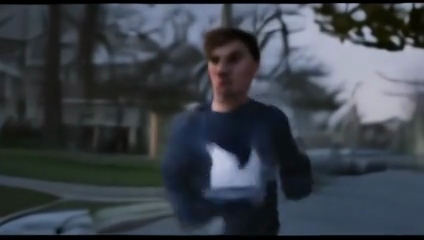}
    \end{minipage}\hfill
    \begin{minipage}{0.161\textwidth}
        \includegraphics[width=\linewidth]{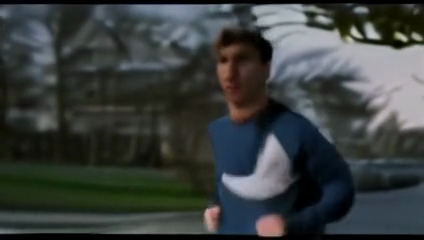}
    \end{minipage}\hfill
    \begin{minipage}{0.161\textwidth}
        \includegraphics[width=\linewidth]{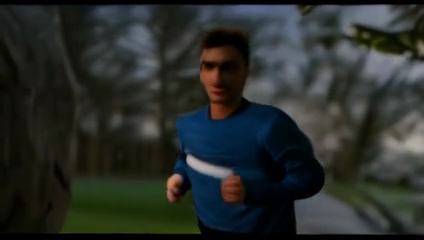}
    \end{minipage}\hfill
    \begin{minipage}{0.161\textwidth}
        \includegraphics[width=\linewidth]{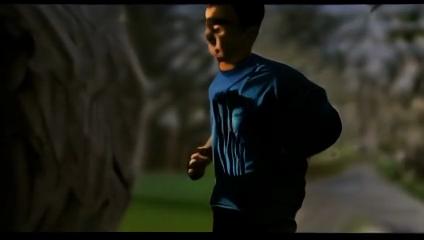}
    \end{minipage}\hfill
    \begin{minipage}{0.161\textwidth}
        \includegraphics[width=\linewidth]{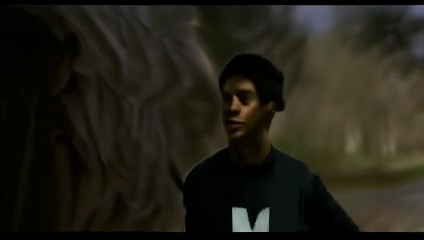}
    \end{minipage}\hfill
    \begin{minipage}{0.161\textwidth}
        \includegraphics[width=\linewidth]{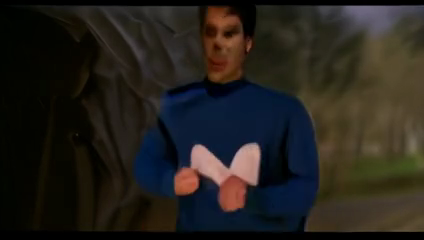}
    \end{minipage}\hfill

    \begin{minipage}{0.02\textwidth} %
        \rotatebox{90}{\small{S-T2V}} %
    \end{minipage}
    \begin{minipage}{0.161\textwidth}
        \includegraphics[width=\linewidth]{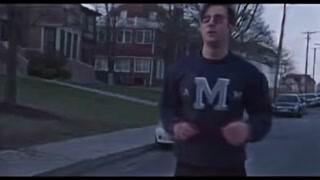}
    \end{minipage}\hfill
    \begin{minipage}{0.161\textwidth}
        \includegraphics[width=\linewidth]{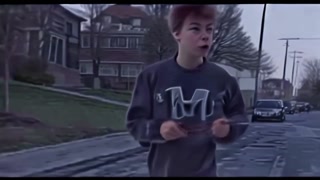}
    \end{minipage}\hfill
    \begin{minipage}{0.161\textwidth}
        \includegraphics[width=\linewidth]{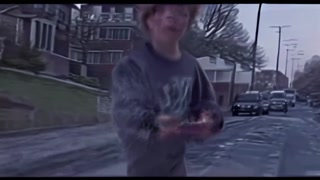}
    \end{minipage}\hfill
    \begin{minipage}{0.161\textwidth}
        \includegraphics[width=\linewidth]{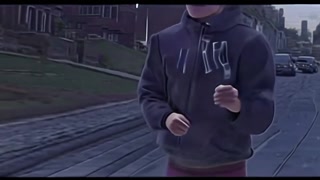}
    \end{minipage}\hfill
    \begin{minipage}{0.161\textwidth}
        \includegraphics[width=\linewidth]{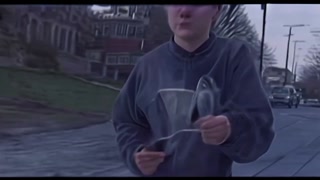}
    \end{minipage}\hfill
    \begin{minipage}{0.161\textwidth}
        \includegraphics[width=\linewidth]{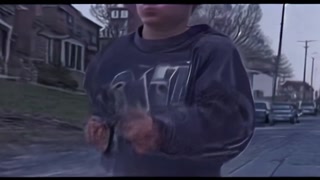}
    \end{minipage}\hfill

    \begin{minipage}{0.02\textwidth} %
        \rotatebox{90}{\small{SVD}} %
    \end{minipage}
    \begin{minipage}{0.161\textwidth}
        \includegraphics[width=\linewidth]{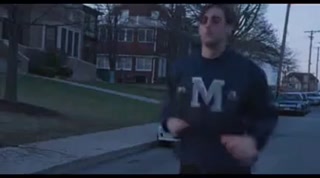}
    \end{minipage}\hfill
    \begin{minipage}{0.161\textwidth}
        \includegraphics[width=\linewidth]{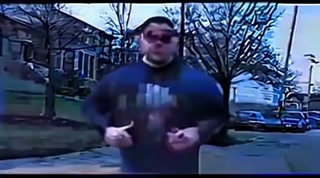}
    \end{minipage}\hfill
    \begin{minipage}{0.161\textwidth}
        \includegraphics[width=\linewidth]{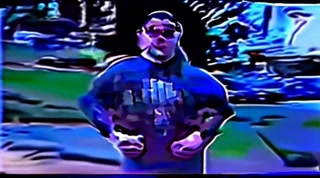}
    \end{minipage}\hfill
    \begin{minipage}{0.161\textwidth}
        \includegraphics[width=\linewidth]{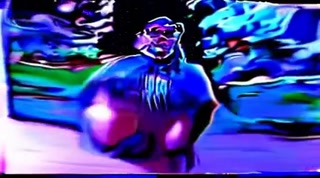}
    \end{minipage}\hfill
    \begin{minipage}{0.161\textwidth}
        \includegraphics[width=\linewidth]{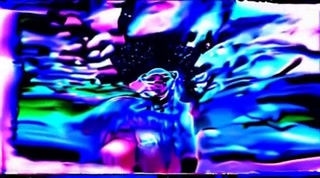}
    \end{minipage}\hfill
    \begin{minipage}{0.161\textwidth}
        \includegraphics[width=\linewidth]{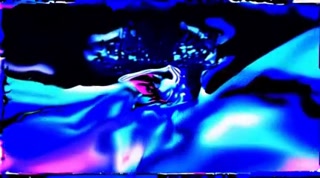}
    \end{minipage}\hfill

    \caption{
    Qualitative comparison of
    PA-\internalmodel (ours), RW-\internalmodel, PA-\opensora-base (ours), RN-\opensora-base, StreamingSVD from StreamingT2V~\cite{henschel2024streamingt2vconsistentdynamicextendable}, SVD-XT from Stable Video Diffusion~\cite{blattmann2023stable}, and FIFO-Diffusion~\cite{kim2024fifo}.
    Frames are evenly sampled from 1 minute long generated video, i.e. at 10, 20, 30, 40, 50, and 60 seconds.
    Our models can autoregressively generate 60-second, 1440-frame videos without quality degradation.
    \todos{FIFO}
    }
    \label{fig:qualitative_appendix}
\end{figure*}

We provide additional quailtative results in~\cref{fig:qualitative_appendix}.

\section{Additional Ablation Study}
\label{sec:add_ablation}
In our project webpage, we show an ablation study on our \textit{Variable Length} design (\cref{sec:method:variable_length}).
We compare Variable Length inference results of PA-\internalmodel models trained with and without Variable Length. 
Without Variable Length training, the second video shows temporal jittering and abrupt scene change at the 1st and 59th seconds. 
This is because the model is not trained to generate the first/last chunk of latent frames to be consistent with the prior chunks.
With Variable Length training, the first video avoids the jittering and abrupt scene change at the 1st and 59th seconds, and the video is temporally smooth. 
Furthermore, Variable Length inference enables the model to generate precisely 1440 frames, whereas without this technique the model would need to discard the noisy chunks remaining in the context window, which correspond to the 1441-1584th frames, when it reaches the 1440th frame. 
Being able to stop the autoregressive video denoising at a precise ending frame allows our model to generate a proper ending to the video, e.g. the woman exits the camera view in the first video, which is not possible without the Variable Length technique.

\begin{table}[]
    \centering
    \begin{tabular}{c|c}
        \toprule
        S & FVD$\downarrow$ \\
        \midrule
        50 & 358.20 \\
        \midrule
        100 & \textbf{339.59} \\
        \midrule
        150 & 399.91 \\
        \bottomrule
    \end{tabular}
    \caption{Ablation on the number of sampling steps $S$ of the PA-\internalmodel model.}
    \label{tab:ablation_s}
\end{table}

Additionally, we ablate the number of sampling steps $S$ of the PA-\internalmodel.
Note that our progressive video denoising can work with arbitrary $S$; when the \textit{chunked frames} technique is used, $S$ only needs to be divisible by $C$.
We compute FVD scores in the same way as described in~\cref{sec:exp:long}.
As shown in~\cref{tab:ablation_s}, further increasing $S$ from 50 to 100 provides marginal benefits despite doubling the inference compute cost, while increasing $S$ to 150 leads to slightly worse results.

\end{document}